\documentclass[10pt,twocolumn,letterpaper]{article}

\usepackage{cvpr}              % To produce the CAMERA-READY version
\usepackage[accsupp]{axessibility}

\definecolor{cvprblue}{rgb}{0.21,0.49,0.74}
\usepackage[pagebackref,breaklinks,colorlinks,allcolors=cvprblue]{hyperref}

\def\paperID{2762} % *** Enter the Paper ID here
\def\confName{CVPR}
\def\confYear{2025}

\title{\vspace{-20pt}Do computer vision foundation models learn \\ the low-level characteristics of the human visual system?}

\author{Yancheng Cai, Fei Yin, Dounia Hammou, Rafal Mantiuk; University of Cambridge, UK\\{{\tt\small yc613@cam.ac.uk}, {\tt\small fy277@cam.ac.uk}, {\tt\small dh706@cam.ac.uk}, {\tt\small mantiuk@gmail.com}}
}

\newcommand{\csdm}{\,cd/m$^2$}

\newcommand{\figref}[1]{Figure~\ref{fig:#1}}
\newcommand{\secref}[1]{Section~\ref{sec:#1}}

\newcommand{\cdms}{\,cd/m$^2$}

\newcommand{\supplementary}{Supplementary}

\renewcommand{\eqref}[1]{eq.~(\ref{eq:#1})}
\newcommand{\norm}[1]{\left\lVert#1\right\rVert}

\DeclareMathOperator*{\argmin}{argmin}

\newcommand{\ind}[1]{\text{#1}}

\begin{document}
\maketitle
\begin{abstract}
Computer vision foundation models, such as DINO or OpenCLIP, are trained in a self-supervised manner on large image datasets. Analogously, substantial evidence suggests that the human visual system (HVS) is influenced by the statistical distribution of colors and patterns in the natural world, characteristics also present in the training data of foundation models. The question we address in this paper is whether foundation models trained on natural images mimic some of the low-level characteristics of the human visual system, such as contrast detection, contrast masking, and contrast constancy. Specifically, we designed a protocol comprising nine test types to evaluate the image encoders of 45 foundation and generative models. Our results indicate that some foundation models (e.g., DINO, DINOv2, and OpenCLIP), share some of the characteristics of human vision, but other models show little resemblance. Foundation models tend to show smaller sensitivity to low contrast and rather irregular responses to contrast across frequencies. The foundation models show the best agreement with human data in terms of contrast masking. Our findings suggest that human vision and computer vision may take both similar and different paths when learning to interpret images of the real world. Overall, while differences remain, foundation models trained on vision tasks start to align with low-level human vision, with DINOv2 showing the closest resemblance. Our code is available on \url{https://github.com/caiyancheng/VFM_HVS_CVPR2025}.

\end{abstract}

\section{Introduction}
\label{sec:intro}
Computer vision foundation models, such as DINO~\cite{caron2021emerging} or OpenCLIP~\cite{ilharco_gabriel_2021_5143773, Radford2021LearningTV}, show exceptional ability to generalize to different tasks and are becoming cornerstones of many computer vision methods. They owe their exceptional performance to self-supervised training on very large image datasets. The human visual system also owes much of its capability to being able to perceive the world, over many years from infancy to childhood \cite{Braddick_Atkinson_2011}. A question arises: if the neural network and the visual system are trained by being exposed to a large number of images of the world, will they share their low-level vision characteristics? If they do, we will know that those low-level characteristics arise naturally and likely reflect the statistics of real-world scenes. If they do not, it means that human low-level vision characteristics are specific to the optical/biological limitations of human vision rather than natural image statistics. Our analysis is meant to shed some light on how the vision, either biological or computational, may develop from observing samples of the world, taking either the same or different routes to accomplish their respective tasks. 

\begin{figure}[t]
  \vspace{-10pt}
  \centering
      \includegraphics[width=\linewidth]{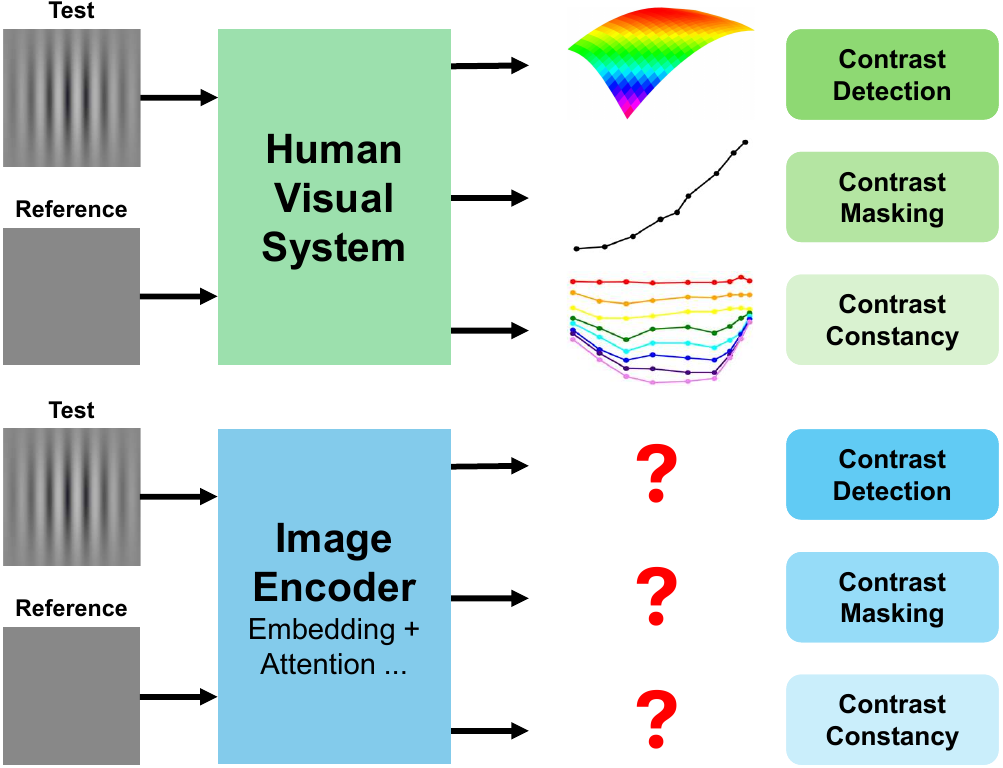}
  \caption{To determine whether image encoders of foundation models exhibit a similar low-level characteristic as human vision, we test them on psychophysical stimuli for which human data is available. We want to test the alignment of contrast encoding between human and computational vision models.}
  \label{fig:dream}
\end{figure}

In particular, we are interested in the characteristics that are well understood and measured in human vision science using psychophysical methods: contrast detection \cite{barten1999contrast}, contrast masking \cite{Legge_Foley_1980} and contrast constancy \cite{georgeson1975contrast}. Contrast detection and contrast masking quantify the ability of the visual system to detect small contrast patterns, either on uniform backgrounds (contrast detection) or on backgrounds with patterns (contrast masking). Contrast detection and masking capture the ``bottlenecks'' of the visual system --- the characteristic that may prevent us from detecting patterns that are too dark or too small. Similarly, cameras used for computer vision are limited by the MTF of the lens, sensor resolution, photon and sensor noise, and we can expect that computer vision methods may need to deal with similar limitations. 

Contrast constancy is the term used in vision science to describe the invariance of the visual system to spatial frequency \cite{georgeson1975contrast} and partially luminance \cite{Kulikowski_1976,Peli_1995}. Georgeson and Sullivan \cite{georgeson1975contrast} showed that the perceived magnitude of the contrast that is well above the detection threshold (supra-threshold) appears to us the same regardless of spatial frequency. This is a very important characteristic as it allows us to see contrast (and therefore objects) the same regardless of the viewing distance; otherwise, the frequencies would change with the viewing distance and hence the contrast appearance. A partial constancy (invariance) is also observed across luminance \cite{Kulikowski_1976,Peli_1995}, though there is a significant deviation from constancy at lower luminance levels, once the visual system needs to rely on the rod vision. The invariance is also an important feature of many computer vision methods. For example in SIFT features \cite{Lowe_2004} have been designed to be invariant to the changes in contrast, brightness, scale, and rotation. In our experiments, we used the supra-threshold contrast matching test to assess whether the models exhibit the characteristic of contrast constancy.

Numerous works on adversarial attacks demonstrated that the classification performance of deep learning models can be greatly degraded by visually inconsequential changes \cite{Goodfellow_Shlens_Szegedy_2014}. At the same time, human vision does not suffer from such adversarial vulnerability \cite{wichmann2023deep}. This is one of the most salient arguments put forward to state that deep architectures are different from human vision. Here, we propose a different methodology to study this question. We consider the deep neural network to be a black box and compare its responses to well-understood and measured characteristics of the human visual system. In particular, we want to check whether the foundation vision models share the same ``bottlenecks'' and invariance properties as the visual system. To achieve this, we test foundation models on basic vision stimuli, such as Gabor patches and band-limited noise, and compare the response of those models with the psychophysical data collected from human observers. 

In summary, our contributions are as follows:
\begin{itemize}
\item We developed a protocol to evaluate the similarity between machine vision models and the human visual system. This protocol includes contrast detection, contrast masking, and contrast constancy, subdivided into nine distinct test types that collectively capture the low-level fundamental characteristics of human vision.
\item We tested the image encoders of 45 foundation and generative models. The results reveal similarities between certain foundation models (\eg, DINOv2 and OpenCLIP) and human vision, particularly in the contrast masking test. However, differences persist across other tests.
\end{itemize}

\section{Related Work}
\label{sec:relat}
Since the advent of deep learning, machine vision models based on foundation models~\cite{ dosovitskiy2020image, oquab2023dinov2, kirillov2023segment} and DNNs~\cite{tian2023multi, lu2023efficient, zhang2025bridgenet, lv2025drkd}  have successfully handled numerous advanced visual tasks. However, researchers have observed that machine vision operates differently from human vision. \cite{geirhos2018imagenet} revealed that standard CNNs trained on ImageNet are strongly biased toward texture recognition rather than shape, which contrasts with human visual patterns. \cite{wichmann2023deep} provided further evidence that deep neural networks (DNNs) differ significantly from the HVS, demonstrating poor robustness in object classification under 3D viewpoint changes and image distortions, and showing vulnerability to adversarial examples, which are rarely problematic for humans. \cite{bowers2023deep} pointed out that DNNs performing well in benchmark tests share little overlap with biological vision mechanisms and fail to account for many findings in psychological studies of human vision. This highlights a clear distinction between machine and human vision, leading to the rise of interest in domain adaptation~\cite{cai2023rethinking} and making networks robust to adversarial attacks~\cite{yin2023generalizable}.

But there is also evidence that the gap between neural network-based machine vision models and human vision is gradually narrowing. \cite{tuli2021convolutional} compared vision transformers (ViT)~\cite{dosovitskiy2020image} and CNNs, finding that ViT not only achieves superior task accuracy but also exhibits weaker inductive biases, with error patterns more consistent with human errors. \cite{geirhos2021partial} discovered that the long-standing robustness gap between humans and CNNs in handling distortions is shrinking. \cite{ghildyal2024foundation, croce2024adversarially} also demonstrated that foundation models like DINO~\cite{caron2021emerging} and CLIP~\cite{radford2021learning} can generate more accurate and robust metrics for low-level perceptual similarity.

Most of the aforementioned studies focus on high-level task performance (\eg, accuracy, consistency, ...), which may not reveal whether computation models suffer from the same bottlenecks and rely on the same invariances as human vision. To that end, \cite{li2022contrast, akbarinia2023contrast} have attempted to reveal CSF characteristics within pretrained architectures by training a head with a contrast discrimination classifier. The problem with this approach is that it introduces a bias by relying on a classifier trained to compare contrast. Such studies also make an incorrect assumption that CSF explains both near-threshold and super-threshold vision, while contrast constancy results (see \secref{sub_SCM}) show that this is not the case.  In contrast, we examine networks' low-level characteristics without additional task-specific training, considering both near-threshold and supra-threshold vision.

\section{Testing framework}
\label{sec:imple}
We first explain the tested models, testing methods, result visualization, and the strategy we used to summarize and quantify our results. 

\subsection{Tested models and testing methodology}

The objective is to evaluate the responses of machine vision foundation models to stimuli commonly used in human vision research~\cite{ashraf2024castlecsf, yancheng2024elaTCSF} and to compare these responses with psychophysical human data. We tested a representative set of 45 models, encompassing the most influential large vision foundation models, including the variants of DINO~\cite{caron2021emerging}, DINOv2~\cite{oquab2023dinov2,darcet2023vitneedreg}, OpenCLIP~\cite{ilharco_gabriel_2021_5143773, Radford2021LearningTV}, SAM~\cite{kirillov2023segment}, SAM-2~\cite{ravi2024sam}, and MAE~\cite{he2022masked}, as well as the encoder used for the latent space of generative model Stable Diffusion (SD-VAE, \cite{rombach2022high}). Additionally, we report the responses of ColorVideoVDP~\cite{mantiuk2024colorvideovdp}, which is an image and video quality metric that explicitly models low-level human vision and acts as a reference for a low-level human vision model. All models and their variants are listed in \figref{All_score_result}.

\begin{figure}[t]
  \centering
  \includegraphics[width=\linewidth]{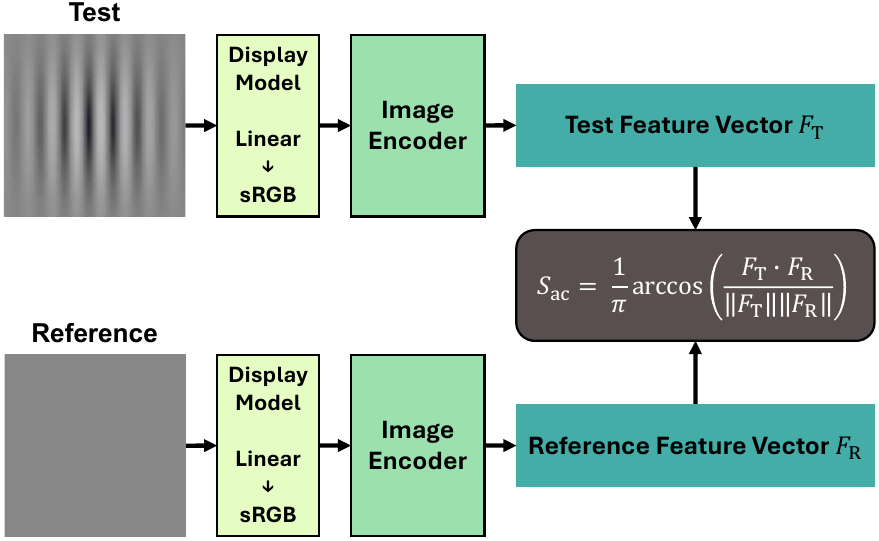}
  \caption{Pipeline for computing $S_\ind{ac}$. Generate test and reference images in the linear luminance space (\csdm), transform them to the sRGB color space using a display model, and input each into the image encoder. Reshape the output features into one-dimensional vectors $F_\ind{T}$ and $F_\ind{R}$, then compute $S_\ind{ac}$.}
  \label{fig:pipeline}
\end{figure}

To test these models, we need to compare pairs of images and assess the ``perceived'' difference between them. We adopt a methodology inspired by 2-alternative-forced-choice (2AFC) psychophysical experiments. For example, for a pattern detection task, a pair of images could be a Gabor patch (test) and a uniform field (reference), as shown on the left of \figref{pipeline}. 
As such patterns are calibrated in physical light units in vision science, we generate these patterns as luminance maps, scaled in physical units of \cdms. These luminance maps are then mapped from the linear space to the sRGB color space using a display model (display peak luminance of 400\cdms) and fed into the image encoder of the foundation model for feature extraction. Note that the sRGB space is almost universally used to represent training datasets and is expected input for the tested encoders. To ensure that models can operate on small contrast values, we modified them to accept floating-point values (instead of 8-bit integers) as input. This was necessary as the quantization artifacts in 8-bit images are often larger than the detection thresholds of human vision.

To investigate whether the distances in the feature space reflect the perceptual detection thresholds and invariances, we experimented with a series of distance measures, including $L_1$ and $L_2$. We found the cosine similarity expressed as a relative angle ($S_\ind{ac}$) yielded results most consistent with the psychophysical data. $S_\ind{ac}$ is defined as:
\begin{equation}
S_\ind{ac} = \frac{1}{\pi}\arccos \left( \frac{F_\ind{T} \cdot F_\ind{R}}{\norm{F_\ind{T}}\norm{F_\ind{R}}}\right)
    \label{eq:sa},
\end{equation}
where $F_\ind{T}$ and $F_\ind{R}$ are the test and reference feature vectors (feature maps reshaped into one dimension), and $\cdot$ denotes the dot product. $S_\ind{ac} = 0$ indicates two input images are equivalent, while $S_\ind{ac}=1$ indicates large differences.

To compare the encoder responses with psychophysical data, we need to be able to map the image sampling frequency into the spatial frequency on the retina. For that, we select the effective resolution of 60 pixels-per-degree, which is typical for modern monitors. We note, however, that the choice of this parameter is arbitrary and the model similarity scores can be shifted by a small multiplier along the spatial frequency axis. The luminance maps are generated at the resolution of 224$\times$224 pixels, corresponding to the size of 3.7$\times$3.7 visual degrees. It roughly aligned with the extend of the human foveal vision, and the stimuli span across multiple receptive fields of human vision and multiple patches/tokens of a foundation model. 

For ColorVideoVDP, it can work directly on linear physical units, so conversion to sRGB was unnecessary. We directly use its quality score instead of $S_\ind{ac}$.  Note that the primary focus of this study is on foundation models, with the ColorVideoVDP metric used solely as a baseline for comparison purposes.

\begin{figure}[t]
  \centering
      \includegraphics[width=\linewidth]{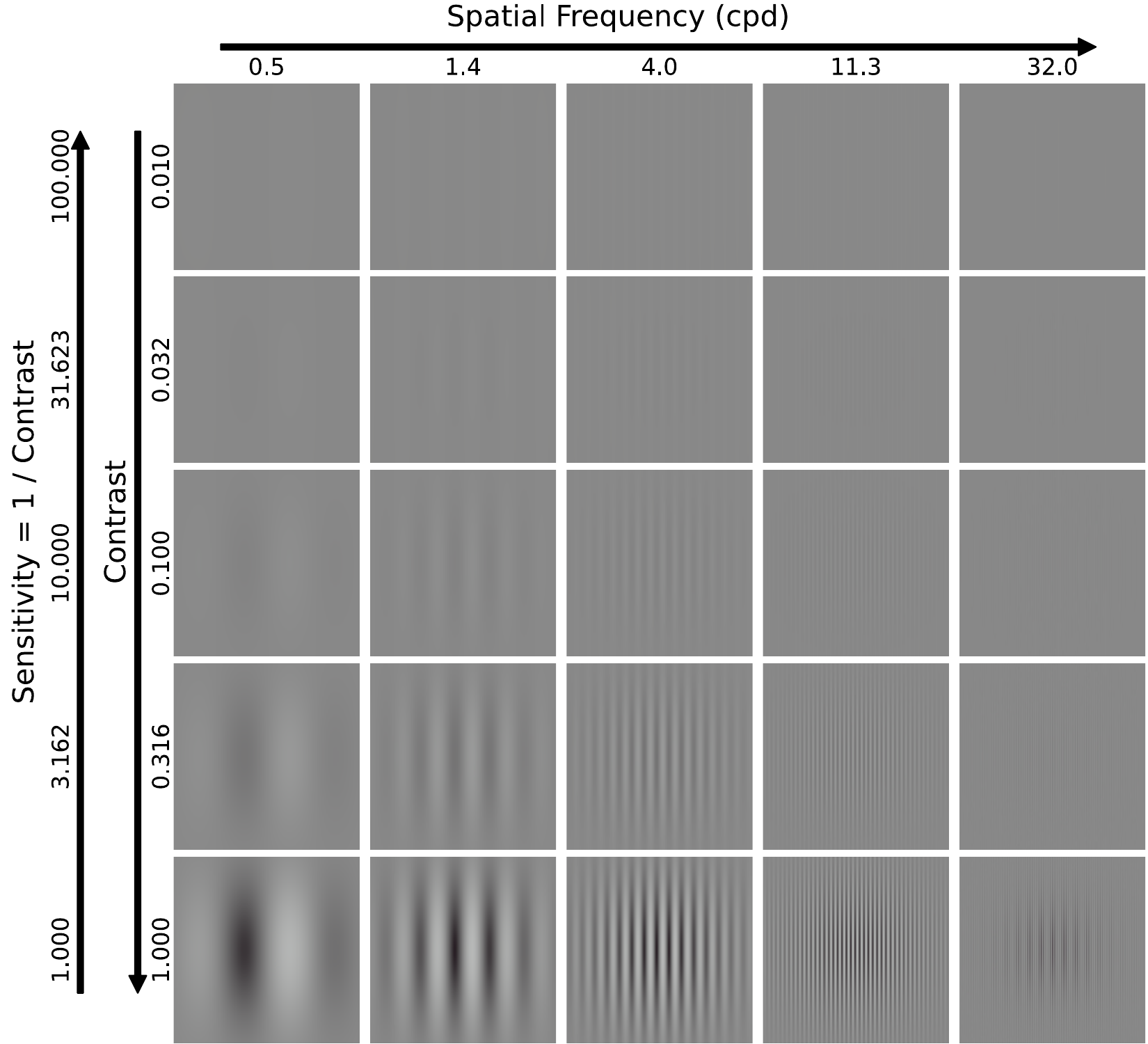}
  \caption{Gabors with different spatial frequencies (x-axis) and contrast (y-axis) used as the test images in the contrast detection tests (\secref{contrast-detection}). ``cpd" denotes cycles per degree. Note that the high-frequency patterns can be rendered with aliasing artifacts on the screen or in print --- those were not present in our tests.}
  \label{fig:gabor_spf_contrast}
\end{figure}

\subsection{Model alignment score}
\label{sec:model-alignment-score}

Most of our results will be represented as contour plots of model responses, providing qualitative interpretation. Here, we explain our measure of model alignment, which provides quantitative scores. 

\begin{figure*}[t]
  \centering
  \vspace{0pt}
      \includegraphics[width=1\textwidth]{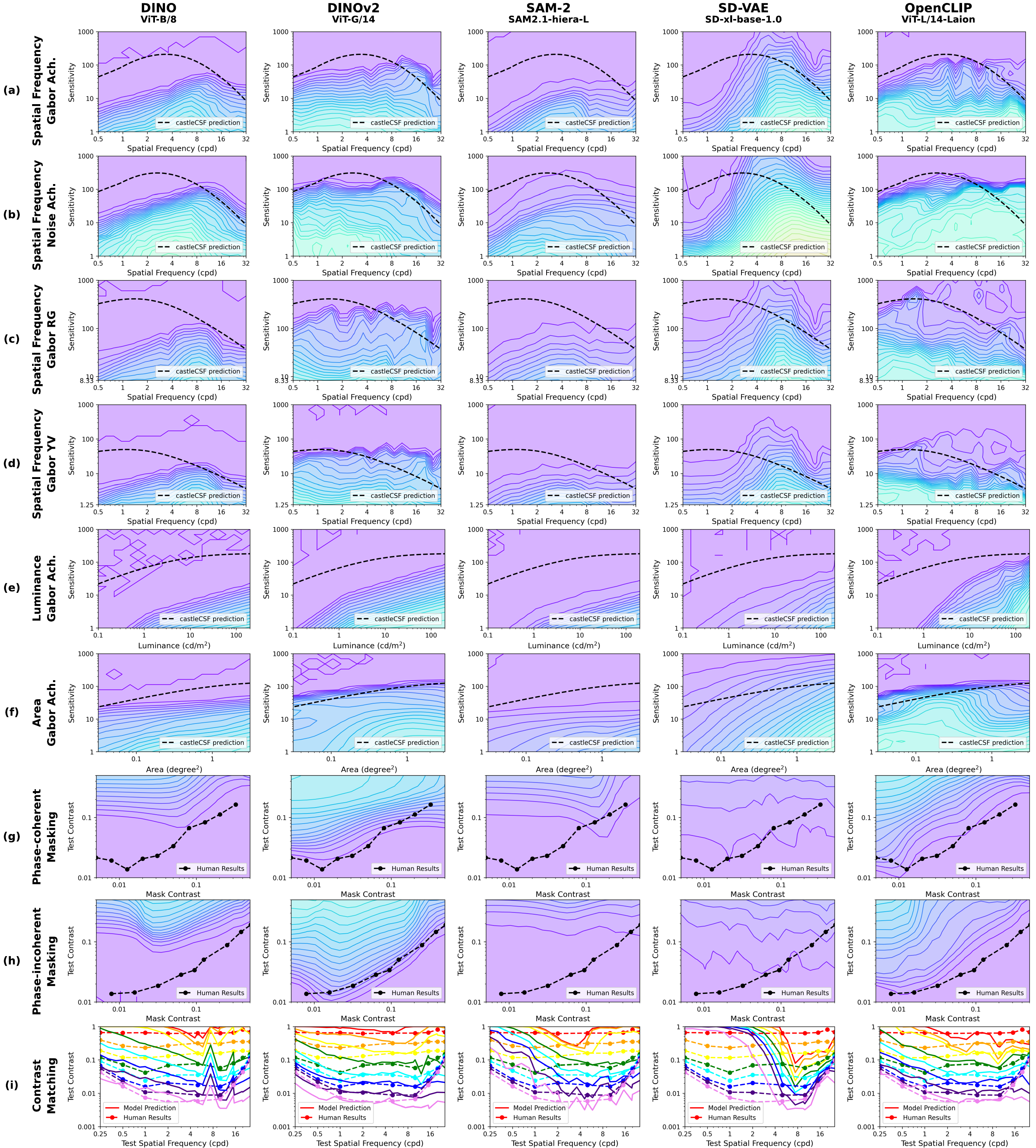}
  \caption{
  Selected representative experimental results. Each row represents a test, and each column corresponds to a model, selected as the best-performing in their original tasks. (a)-(f): contour plots of contrast detection $S_\ind{ac}$, with the ground truth castleCSF~\cite{ashraf2024castlecsf}. (g),(h): contour plots of contrast masking $S_\ind{ac}$, with the ground truth from~\cite{foley1994human} and~\cite{gegenfurtner1992contrast}, respectively. Different colored solid lines representing different $S_{\mathrm{ac}}$ values, where purple indicates the minimum difference ($S_{\mathrm{ac}}\to0$). (i): results from the contrast matching experiment, where different colors represent different $C_r$ values. The dashed lines are human results~\cite{georgeson1975contrast}, while the solid lines are model-predicted results from Equation~\ref{eq:cm_cos}. Results for all models are in the supplementary material.}
  \label{fig:contour-main}
  \label{fig:Contour_1}
\end{figure*}

As an example, we take the leftmost contour plot in row (a) of \figref{contour-main}. Each point on the contour plot corresponds to $S_\ind{ac}$ between the test image as shown in \figref{gabor_spf_contrast} and a uniform field of the same mean luminance. The dashed line represents human contrast detection data, predicted with castleCSF \cite{ashraf2024castlecsf}. A well-aligned model should show one of the contour lines that follows the dashed castleCSF line. We cannot directly use the $S_\ind{ac}$ values along the dashed line as the measure of alignment because some models result in $S_\ind{ac}=0$ for most points near the detection threshold (detect no difference). Therefore, instead, we rely on the measure of change in $S_\ind{ac}$ in the neighborhood of the dashed line. 

For a well-aligned model, the perceived differences in the neighborhood of the detection threshold (dashed line) should increase as the contrast increases (the sensitivity decreases), Furthermore, the values should be similar along the dashed line. The measure these two properties, we sample the $S_\ind{ac}$ values for the points that are shifted in contrast (vertical direction) from the dashed line by a multiplier $m$, where $0.5{\leq}m{\leq}2$ (note the logarithmic scale in the contour plot in \figref{contour-main}). We collect such data for multiple frequencies (or other dimensions) along the dashed line and calculate the Spearman rank order correlation between the multipliers $m$ and the $S_\ind{ac}$ values. If the properties mentioned above are preserved, the correlation coefficient value $r_s$ should be close to 1. 

The above strategy is used for all contrast detection and contrast masking experiments. For the contrast matching experiment, we use the root mean squared error (RMSE) between the model and human matching data, expressed as the logarithm of contrast. 

\section{Experiments}
\label{sec:exper}
\begin{figure*}[t]
  \centering
      \includegraphics[width=\textwidth]{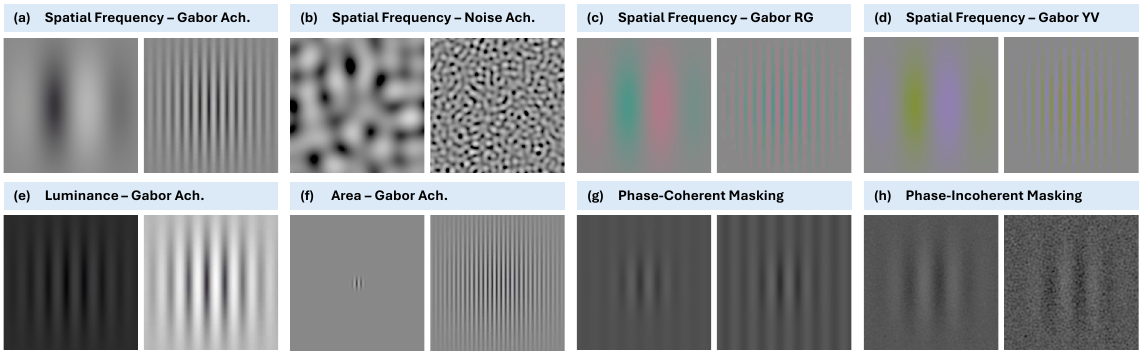}
  \caption{Examples of test images for contrast detection and contrast masking. The detailed explanation of the stimuli can be found in the \supplementary{}. ``Ach." denotes achromatic. }
  \label{fig:test_examples}
  \label{fig:stimuli}
\end{figure*}

\begin{table*}[t]
\begin{center}
\caption{
Key parameters for all our tests. Note that ``Radius" does not apply when the pattern is not a Gabor.
}
\label{tab:params}
\setlength{\tabcolsep}{0.7mm}
\footnotesize
{\scalebox{1.05}{
\begin{tabular}{c|c|c|c|c}
\toprule
Test & Spatial Frequency (cpd) & Luminance (\csdm{}) & Radius (degree) & Contrast \\ \hline\hline
Spatial Frequency - Gabor   Achromatic & 0.5 - 32 & 100 & 1 & 0.001 - 1 \\ \hline
Spatial Frequency - Noise   Achromatic & 0.5 - 32 & 100 & - & 0.001 - 1 \\ \hline
Spatial Frequency - Gabor RG & 0.5 - 32 & 100 & 1 & 0.001 - 0.12 \\ \hline
Spatial Frequency - Gabor YV & 0.5 - 32 & 100 & 1 & 0.001 - 0.8 \\ \hline
Luminance - Gabor Achromatic & 2 & 0.1 - 200 & 1 & 0.001 - 1 \\ \hline
Area - Gabor Achromatic & 8 & 100 & 0.1 - 1 & 0.001 - 1 \\ \hline
Phase-Coherent Masking & 2 (mask) / 2 (test) & 32 & 0.5 (test) & 0.005 - 0.5 (mask) / 0.01 - 0.5 (test) \\ \hline
Phase-Incoherent Masking & 0 - 12 (mask) / 1.2 (test) & 37 & 0.8 (test) & 0.005 - 0.5 (mask) / 0.01 - 0.5 (test) \\ \hline
Contrast Matching & 5 (reference) / 0.25 - 25 (test) & 10 & - & 0.005 - 0.629 (reference) \\ \bottomrule
\end{tabular}
}}
\end{center}
\end{table*}

\begin{figure}[t]
  \centering
      \includegraphics[width=\linewidth]{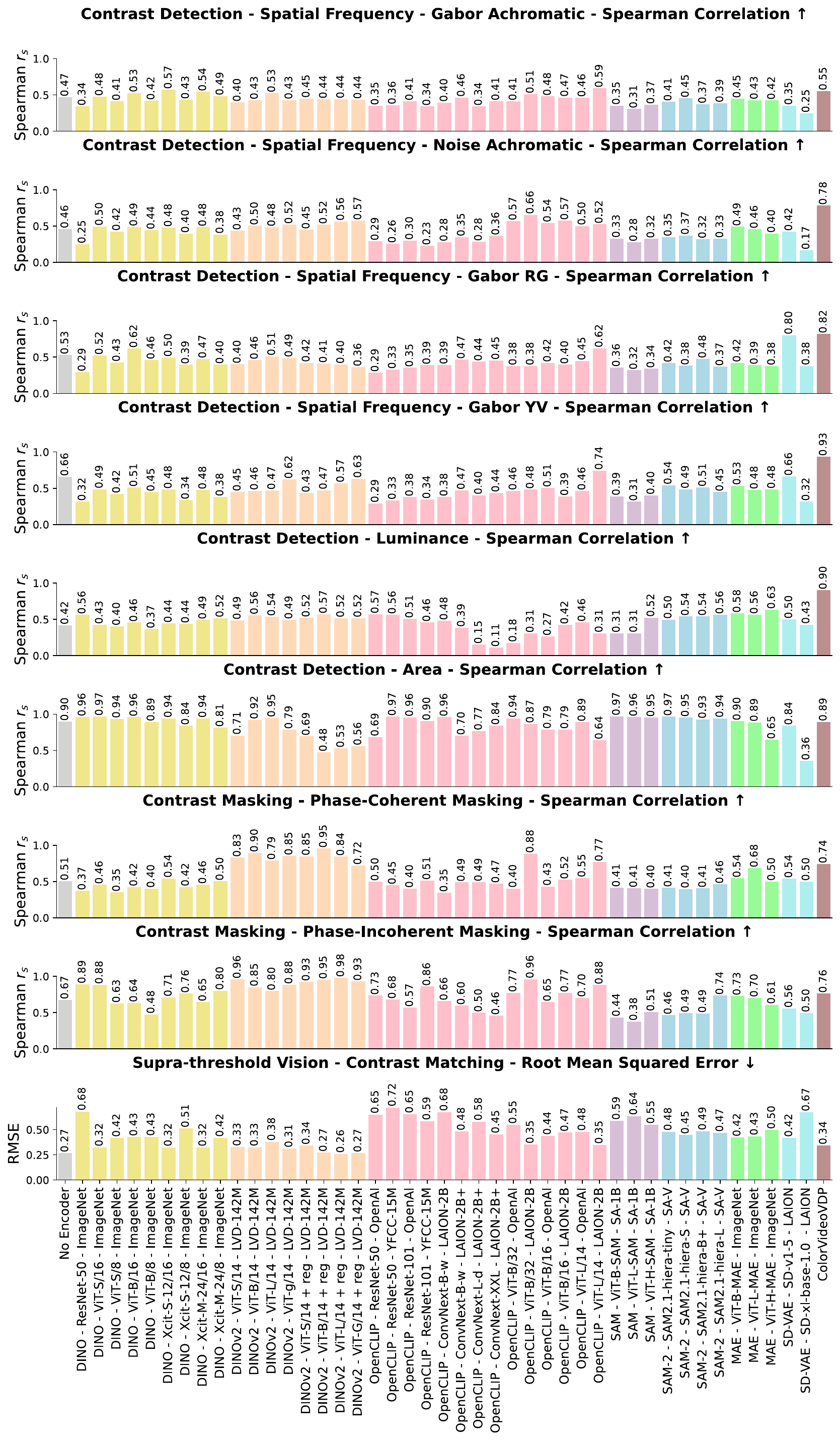}
  \caption{The quantified similarity error between all 45 models and HVS under 9 different tests. For the Contrast Detection and Contrast Masking tasks, Spearman Correlation was used as the metric, with higher values (closer to 1) indicating greater similarity to human vision. For the Supra-threshold Contrast Matching task, RMSE was used as the metric, with lower values (closer to 0) indicating better similarity.}
  \label{fig:All_score_result}
\end{figure}

\subsection{Contrast detection and CSF}
\label{sec:sub_CD}
\label{sec:contrast-detection}

We begin by testing the foundation models' ability to detect low-contrast (near-threshold) patterns (\eg, Gabor patches, band-limited noise) and compare their performance with the human data. As the reference human data, we rely on the caslteCSF \cite{ashraf2024castlecsf}, which is the recent contrast sensitivity function, modeling contrast detection of both achromatic and chromatic patterns.

\paragraph{Spatial Frequency}

Contrast sensitivity of the human eye is typically associated with the variation across the spatial frequency. The visual system exhibits a band-pass characteristic, with a peak sensitivity between 2 and 4 cycles per degree (cpd), depending on the luminance and other parameters of the stimulus. The lower sensitivity of the visual system at lower frequencies is associated with the mechanism of lateral inhibition \cite{barten1999contrast}, which helps to reduce the influence of (low-frequency) illumination on the perceived images. Such invariance to illumination is a desirable property if the goal is to recognize objects regardless of illumination conditions. The drop in sensitivity at high frequencies is associated with the limitations of eye optics (achromatic contrast) and cone density (chromatic contrast). Here we want to test whether the computer-vision foundation models pick up similar traits when trained on natural images. 

To test encoder responses across frequencies, we generated a 2D array of image pairs, in which the reference image had uniform luminance, and the test image contained a Gabor patch, as shown in \figref{gabor_spf_contrast} (refer to \supplementary{} for the visualization of other stimuli). We generate such Gabors for achromatic (see \figref{stimuli}-a) and chromatic modulation (see \figref{stimuli}-c,d). We also tested band-limited noise (see \figref{stimuli}-b). The test patterns had fixed size but varying spatial frequency (x-axis) and contrast (y-axis). Because most of the contrast detection data are plotted as the function of sensitivity, we follow this convention and plot the sensitivity, which is the inverse of the contrast. This corresponds to the reversal of the axis on the logarithmic plots, which we use in our analysis. The other parameters of the stimuli are listed in Table~\ref{tab:params}. Although we tested multiple variants of each foundation model, here we show only the variant with the highest complexity and best performance on its original high-level tasks. The contour plots for other variants can be found in the \supplementary{}.  

The contour plots of foundation model responses are shown together with a contrast sensitivity function (castleCSF \cite{ashraf2024castlecsf}) in rows (a)--(d) of \figref{contour-main}. If the foundation models had the same contrast detection characteristics as the human eye, we would expect the smallest difference (the lowest $S_\ind{ac}$) contour line to follow the black dashed curve of castleCSF. This is not the case for any of the tested foundation models. Some models, and in particular DINOv2 and SD-VAE, show overall band-pass characteristics, in particular for noise and achromatic Gabors. SD-VAE has a drop of sensitivity at lower frequencies much larger than that of the visual system. The responses to chromatic patterns (rows (c) and (d)) tend to be less regular than to achromatic patterns and lack the shift toward lower frequencies, which is observed in the human data. Both OpenCLIP and SAM-2 show a very inconsistent response across the spatial frequencies. 

From the data, we can conclude that foundation models do not follow the sensitivity pattern of the visual system, but some may show a band-pass characteristic that is associated with the CSF. Many models show lower sensitivity at lower frequencies, which may indicate that those models obtained some invariance to (low-frequency) illumination through training. 

\paragraph{Luminance}

The sensitivity of the human eye increases with the luminance. In dim light, human contrast sensitivity increases proportionally to the square root of retinal illuminance, following the DeVries-Rose law. Conversely, in bright light, sensitivity follows Weber’s law, remaining independent of illuminance~\cite{rovamo1995neural}. In this experiment, we want to test whether the computer vision models lose sensitivity at lower luminance levels. Such a loss could be justified by camera noise, which increases in terms of contrast as the light intensity decreases~\cite{Aguerrebere_2013}. 

To produce contour plots, we generated a 2D array of image pairs in a similar manner as for spatial frequency variations in the section above, but instead of varying spatial frequency, we varied luminance, as shown in \figref{stimuli}-e. As a reminder, we did not pass the absolute luminance values directly to each model but instead converted them to display-encoded (gamma-corrected) sRGB space --- the colour space used for training datasets (see \figref{pipeline}). Such an encoding partially compensates for perceptual non-uniformity of luminance. 

The results, shown in row (e) of \figref{contour-main}, indicate a systematic drop in sensitivity with luminance for all tested models. The drop in sensitivity of those models is faster than that observed in the human data, in particular for OpenCLIP.

We can conclude that the models trained on sufficiently large datasets mimic the sensitivity of the visual system to luminance but with a faster drop in sensitivity. This, however, may be the result of using sRGB representation for training datasets, which were presumably mostly well-exposed and contained relatively little information in darker image regions.

\paragraph{Area}
Stimulus area (size) significantly affects sensitivity, as larger stimuli activate more retinal cells. Sensitivity increases with area up to the saturation point, denoted as the critical area~\cite{Rovamo_1993}. The response to stimuli of different sizes tests the model's ability to pool information across the visual field. 

In this experiment, we vary the size of the Gaussian envelope limiting stimulus size to observe its effect on model responses. We follow the same procedure as in the two previous sections and generate pairs with a uniform field and a Gabor patch (see \figref{stimuli}-f). The parameters of the Gabor patch are listed in Table~\ref{tab:params}. 

The results, shown in row (f) of \figref{contour-main}, indicate that most models show summation across the area (the $S_\ind{ac}$ values increase with the area) and the rate of the increase varies across the models; DINO and SAM-2 have a smaller increase, SD-VAR has a higher increase of $S_\ind{ac}$ than the human data, and only DINOv2 roughly matches human performance. OpenCLIP shows no consistent patterns. We can conclude that many models show spatial pooling characteristic that shares the trend observed in human data, though the actual slope of the increase is typically different.

\subsection{Contrast masking}
\label{sec:sub_CM}
Contrast masking explains the decreased visibility of a signal (test) due to the presence of a supra-threshold background (mask). The masking function defines the relationship between the threshold test contrast required for signal detection and the mask contrast. Put simply, a pattern is more difficult to detect in the presence of another pattern of similar spatial frequency and orientation. A typical masking characteristic of the visual system is shown in rows (g) and (h) of \figref{contour-main}. It consists of a relatively shallow segment at low mask contrast (near the detection threshold), with the slope increasing for high mask contrast. The shape of the curve is influenced by the specific properties of the mask and test signals~\cite{daly1992visible, watson1997model}. We will consider the case in which the masker is a sinusoidal grating of the same frequency as the test pattern (phase-coherent masking, row (g)) and when the masker is noise (phase-incoherent masking, row (h)) \cite{daly1992visible, gegenfurtner1992contrast, foley1994human}, as shown in~\figref{stimuli}-g,h.

First, we consider the fundamental form of \emph{phase-coherent masking} in which the masker image (reference in \figref{pipeline}) is a sinusoidal grating and the test image is the masker plus a Gabor patch of the same frequency and phase as the masker (see \figref{test_examples}g). Contrast masking data, shown as the dashed black curve in row (g) of \figref{contour-main}, shows the smallest contrast of the test Gabor that is detectable in the presence of the masker of a given contrast. As the contrast of the masker increases, the smallest detectable contrast of the test also needs to increase. However, such an increase starts only for a masker that has the contrast sufficiently high to be detected. If the contrast of the masker is near the detection threshold, we can observe a dipper effect --- the contrast detection is facilitated by a masker~\cite{foley1994human, watson1987efficiency, wilson1979four}. Such an effect can be only observed in phase coherent masking. 

As an example of \emph{phase-incoherent masking}, we will consider a masker with band-limited noise and a test with a Gabor patch --- see \figref{test_examples}h. The human detection thresholds for such masking patterns are similar to those for phase-coherent masking, except that the dipper effect disappears \cite{van1988effects}. 

The differences predicted by DINOv2 and OpenCLIP are surprisingly well-aligned with the human contrast masking data --- their responses roughly match the slopes of the human data. The alignment is stronger for the phase-incoherent masking. This is particularly notable for OpenCLIP, which did not show any consistent trends for contrast detection. Other models do not show strong alignment with the human data. 

Overall, computational models are better aligned with human data for contrast masking than for contrast detection. One possible explanation is that the signals that induce contrast masking are plentiful in natural images, but contrast detection stimuli, which involve barely noticeable patterns on uniform backgrounds, are rare. Therefore, computational models are more likely to pick up the characteristic that is well represented in the training datasets. 

\subsection{Supra-threshold contrast matching}
\label{sec:sub_SCM}

While contrast detection and contrast masking explain the just detectable (near-threshold) contrast, most of the vision tasks, such as detection or recognition, involve well-visible (supra-threshold) contrast. Supra-threshold human vision has been studied in contrast-matching experiments in which the magnitude of one contrast is visually matched to the magnitude of another contrast of a different frequency \cite{georgeson1975contrast} or luminance \cite{Kulikowski_1976,Peli_1995}. One of the most significant findings of those studies is \emph{contrast constancy} \cite{georgeson1975contrast} --- the ability of the visual system to match physical contrast across frequencies and luminance levels. The results of the seminal study of Georgeson and Sullivan \cite{georgeson1975contrast} on matching contrast across frequencies are shown as dashed lines in row (i) of \figref{contour-main}. At small contrast, the dashed lines show a band-pass shape that follows the contrast sensitivity function. However, as the contrast is increased, the lines become flat showing little influence of frequency on contrast perception. This is an important property that lets us see objects to have the same appearance regardless of the viewing distance. Such a scale invariance is also important for neural networks that are tasked to detect or recognize objects regardless of their size. 

We followed the experimental setup from~\cite{georgeson1975contrast}, where the reference was a 5\,cpd, 10\csdm{} sinusoidal grating, presented at eight distinct contrast levels $c_\ind{r}$. The test stimulus had the same luminance but a different spatial frequency $\rho_\ind{t}$. In~\cite{georgeson1975contrast}, observers adjusted the test stimulus contrast $c_\ind{t}$ until its apparent contrast matched that of the reference (contrast matching). In our experiments, we match contrast encodings of sinusoidal gratings: the ($S_\ind{ac}()$, \eqref{sa}), between a feature vector of a sinusoidal grating, $F(\rho, c)$ of frequency $\rho$ and contrast $c$, and a uniform field, $U=F(\rho_\ind{t}, 0)$. We find the test contrast $c_\ind{t}$ that minimizes the expression:
% \begin{equation}
%     \argmin_{c_\ind{t}} \left(\frac{S_\ind{ac}(F(\rho_\ind{r}, c_\ind{r}), U)}{S_\ind{ac}(F(\rho_\ind{r}, 1), U)} - \frac{S_\ind{ac}(F(\rho_\ind{t}, c_\ind{t}), U)}{S_\ind{ac}(F(\rho_\ind{t}, 1), U)}
%   \right)^2\,,
%   \label{eq:cm_cos}
% \end{equation}
\begin{equation}
    \argmin_{c_\ind{t}} \left(S_\ind{ac}(F(\rho_\ind{r}, c_\ind{r}), U) - S_\ind{ac}(F(\rho_\ind{t}, c_\ind{t}), U)
  \right)^2\,,
  \label{eq:cm_cos}
\end{equation}
where the reference frequency $\rho_\ind{r}=5$\,cpd. The denominators in the expression are used to normalize contrast across frequencies. We experimented with other contrast encodings, including a direct comparison of feature vectors, but the formula above resulted in the best contrast constancy properties across the models. 

The matching contrast predictions for foundation models are visualized as continuous lines in row (i) of \figref{Contour_1}. The plots show that only DINOv2 and OpenCLIP roughly follow the dashed contrast constancy lines. Both models show less attenuation (more constancy) at the highest spatial frequencies, which could be advantageous when the model needs to work with small-scale features. Both models show attenuation of low frequencies (below 1\,cpd), suggesting worse contrast constancy in that frequency range. Other models, including DINO and SAM-2, suffer from large instability across frequencies, or very heavy attenuation of low frequencies in the case of SD-VAE. To conclude, we can observe only partial contrast constancy for selected models. 

\subsection{Model alignment scores}

As the analysis of all 45 variants of the models is infeasible in the scope of this paper, we prepared quantitative results according to the method explained in \secref{model-alignment-score} and summarized them in \figref{All_score_result}. Those let us make three observations:

%The primary focus of this analysis is the visualization in~\figref{contour-main}, which reveals far more detailed insights than the quantitative results alone. However, due to space constraints, including all visualizations in the main text is infeasible. Thus, we present the quantitative results (Model Alignment Scores, defined in Section 3.2) in~\figref{All_score_result}.

First, ColorVideoVDP, which is a visual metric that models human low-level vision, is better aligned with the human data in almost all contrast detection tasks and in terms of contrast constancy, as expected. However, certain variants of OpenCLIP and DINOv2 can match or surpass the ColorVideoVDP alignment in terms of contrast masking. 
%although certain OpenCLIP and DINOv2 variants achieve higher alignment scores on phase-coherent masking tasks, foundation models generally do not surpass the alignment scores of the perception-based image quality metric, ColorVideoVDP, across most dimensions.
Second, the alignment scores of different foundation model variants (\eg, OpenCLIP) show significant variations and alignment scores appear unrelated to the complexity of the variants or their performance on higher-level tasks. %This suggests potential intrinsic randomness within these models, indicating an unreliable capture of low-level human visual characteristics.
Finally, DINOv2 variants, which have been trained to solve vision tasks, show the greatest alignment with the human data among all foundation models. %This could be an indicator that the training used in those models resulted in similar contrast encoding as that found in human vision.  

%This may indicate that future computer vision foundation models will be even more aligned with the low-level contrast encoding of the human visual system. 

%, indicating a current trend toward closer alignment with the human visual system.

% \subsection{Further Validation: Classification Masking}
% \begin{figure}[t]
%   \centering
%       \includegraphics[width=\linewidth]{images/Application_Noise_Masking_2.png}
%   \caption{Image Classification Noise Masking. The upper half presents image examples under varying mask noise contrast levels. The lower half shows accuracy curves across 20 generated mask contrast levels for four DINOv2 architectures.}
%   \label{fig:Application_Noise_Masking}
% \end{figure}

% \RM{This section needs to be updated or removed, as we discussed.}

% The tests above focus on perceptual experiments. To further validate the impact of HVS-model similarity on computer vision tasks, we evaluated noise masking in the image classification task. Specifically, we sampled 5 images from each of the 1000 classes in ImageNet’s validation dataset, adding 20 different noise masking contrasts to each image (using the noise mask in~\figref{test_examples} h), generating a total of 100k images. We then tested four full DINOv2 classification architectures. The results (\figref{Application_Noise_Masking}) clearly show stable classification accuracy at low masking contrasts, followed by a sharp decline at high masking contrasts, consistent with the fundamental characteristics of HVS. 

\section{Conclusions}

If we believe that the goal of both biological and computational low-level vision is to efficiently encode visual information, we can expect that computational models trained on large natural image datasets will share similarities with human vision. In this work, we find that selected computational models, e.g., variants of DINOv2 and OpenCLIP, show surprisingly high alignment with supra-threshold human contrast masking and contrast matching data, but little alignment with the near-threshold contrast detection. This means that computation models do not have the same ``bottlenecks'' as human vision, but, through training, they attain invariance and efficient contrast coding that resembles that of the visual system. We hope that our testing protocol with basic psychophysical stimuli will provide a useful tool for examining future computational models of vision.

{
    \small
    \bibliographystyle{ieeenat_fullname}
    \bibliography{LVM_sythestic_test}
}


\end{document}

\maketitle

This supplementary material provides detailed information on the following: (1) the formulas and examples of the experimental stimuli tested; (2) some further practical implications of our work; and (3) the detailed formula for model alignment scores, along with the model alignment scores for all models across all tests. 

Please open \verb+webpage/index.html+ for the complete set of the results. 

\section{Test images}

The achromatic Gabor patches used for tests are defined as:
\begin{equation}
  G(x, y)=L_{b}\left(1+c\,\sin \left(2 \pi\frac{\rho\,x}{\text{ppd}}\right) e^{\left(-\frac{x^2+y^{2}}{2 \text{ppd}^{2} R^{2}}\right)} \right),
\end{equation}
where $L_{b}$ denotes the background/mean luminance in \cdms{}{}, $c$ represents the contrast, $R$ is the Gabor radius in visual degree, and $\rho$ is the spatial frequency in cycles-per-degree (cpd). $x$ and $y$ represent the image coordinates, where $x\in[-\frac{W}{2}, \frac{W}{2}]$ and $y\in[-\frac{H}{2}, \frac{H}{2}]$; $W=224$ and $H=224$ denote the image width and height, respectively.

To generate chromatic (RG and YV) Gabor stimuli, a single-channel Gabor patch is first created, the color direction is set, then converted to DKL color space, transformed to LMS color space, and finally converted back to RGB to check for gamut constraints. Note that the RGB channel values here are still represented in \cdms{} units.

The luminance values, $G(x,y)$, were converted to RGB values using the sRGB display model, assuming the peak luminance of 400\cdms. The pixels-per-degree (ppd) was set to 60, which approximates the ppd value for a typical human observer viewing an Ultra HD display ($3840 \times 2160$). The resolution of all test and reference images was set to $224 \times 224$. Except for Supra-threshold Contrast Matching (Section~\ref{sec:subsec_supra_CM}), the references for all other eight experiments are uniform achromatic images with a luminance of 100\cdms.

\subsection{Contrast detection}

\begin{figure*}[ht]
  \centering
      \includegraphics[width=\linewidth]{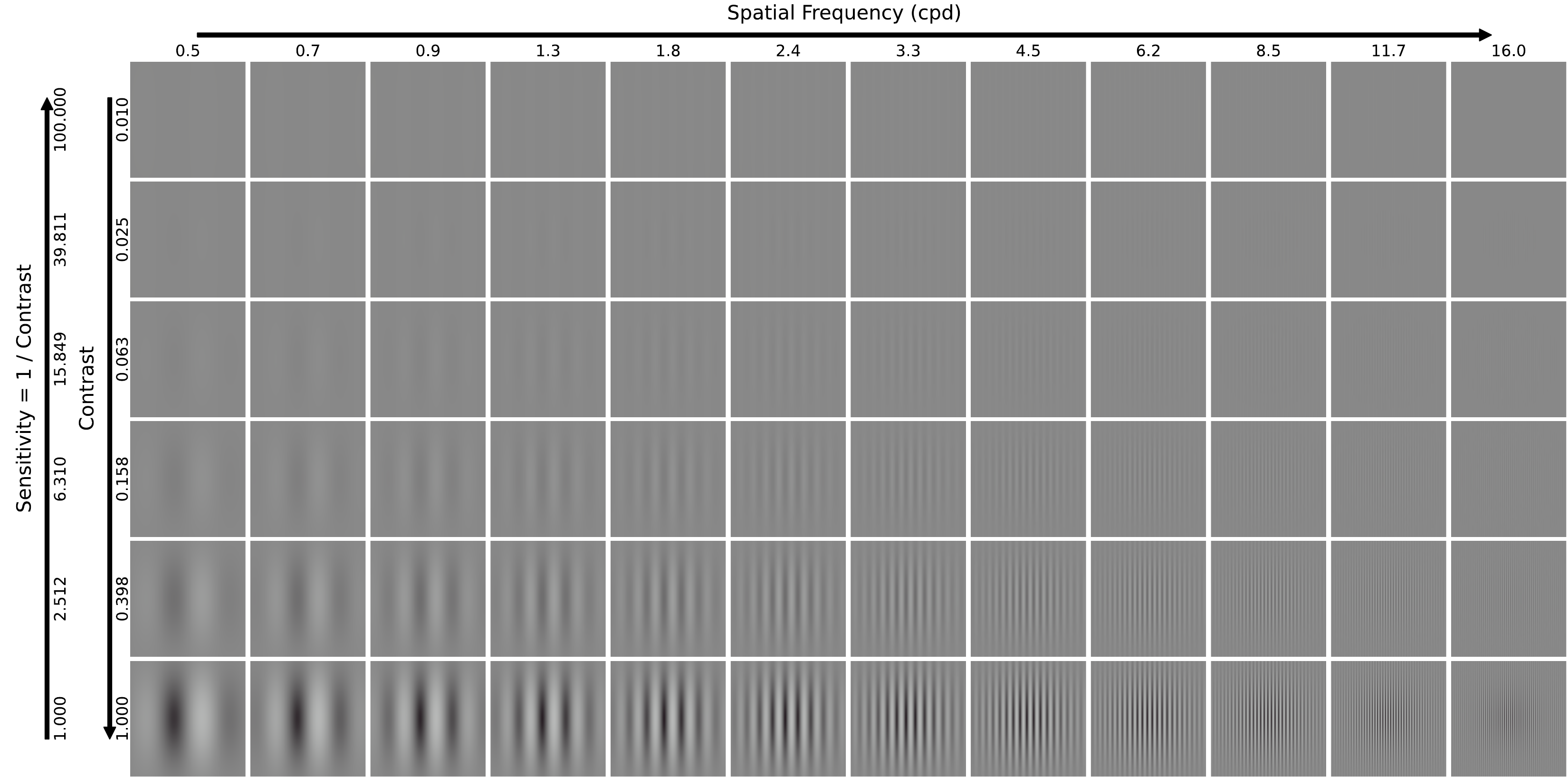}
  \caption{Achromatic Gabors with different spatial frequencies (x-axis) and contrast (y-axis) used as the test images in the contrast detection tests. ``cpd" denotes cycles per degree. High-frequency patterns may introduce aliasing artifacts on screens or prints, so we display up to 16\,cpd here (no such artifacts were present in our tests). Observations indicate that the human eye is indeed most sensitive to achromatic Gabor patterns with spatial frequencies around 2--4\,cpd.}
  \label{fig:gabor_spf_contrast_sup}
\end{figure*}

\paragraph{Spatial Frequency - Achromatic - Gabor}
The radius was set to 1$^\circ$, and the background luminance was 100\cdms{}. Test examples are shown in~\figref{gabor_spf_contrast_sup}.

\begin{figure*}[ht]
  \centering
      \includegraphics[width=\linewidth]{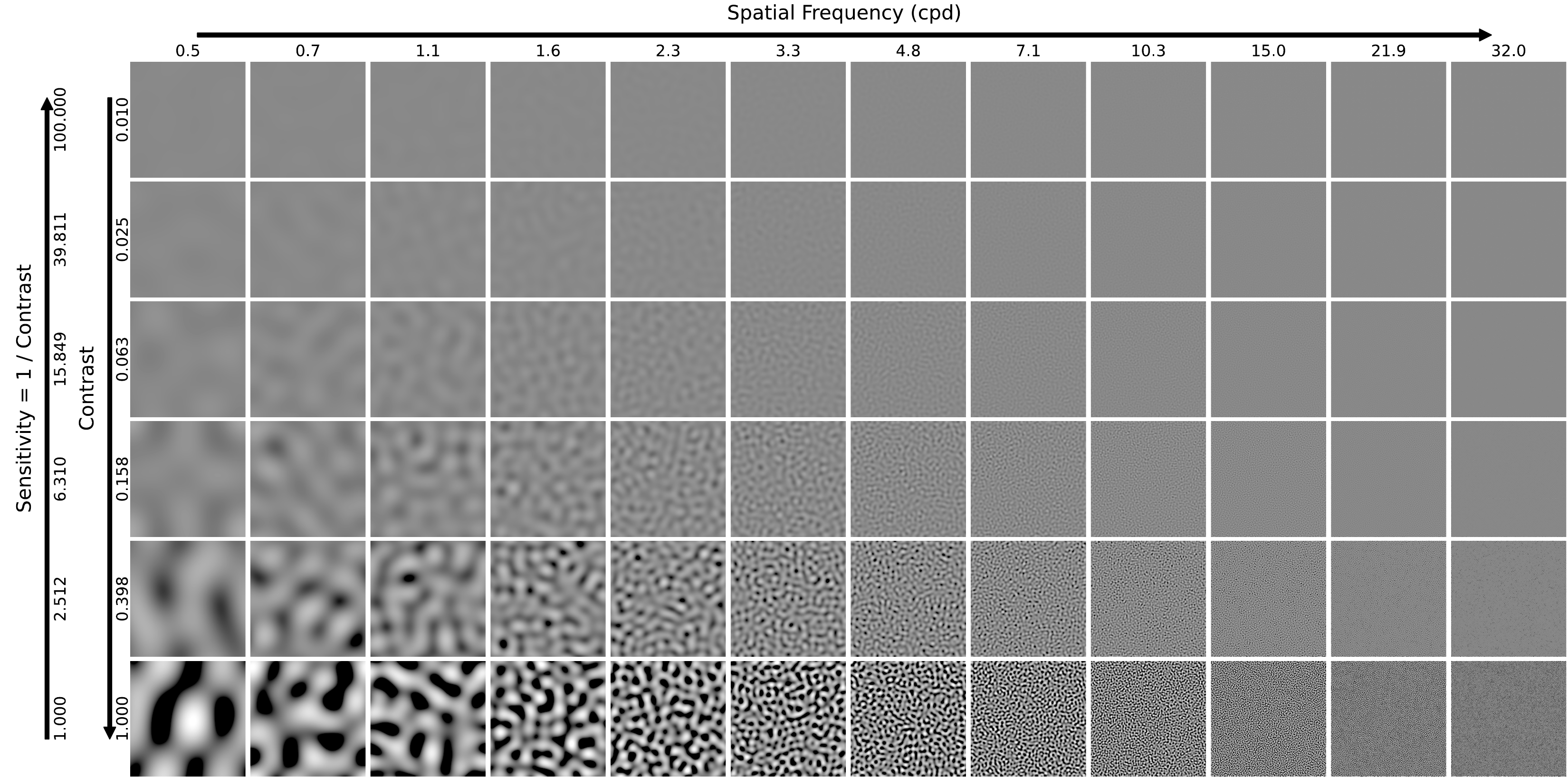}
  \caption{Achromatic band-limited noise signals with different spatial frequencies (x-axis) and contrast (y-axis) used as the test images in the contrast detection tests. Human observers were most sensitive to frequencies in the 2--4\,cpd range.}
  \label{fig:band_limited_noise_sup}
\end{figure*}

\paragraph{Spatial Frequency - Achromatic - Band-limited Noise}
The background luminance was 100\csdm. Test examples are shown in~\figref{band_limited_noise_sup}.

\begin{figure*}[ht]
  \centering
      \includegraphics[width=\linewidth]{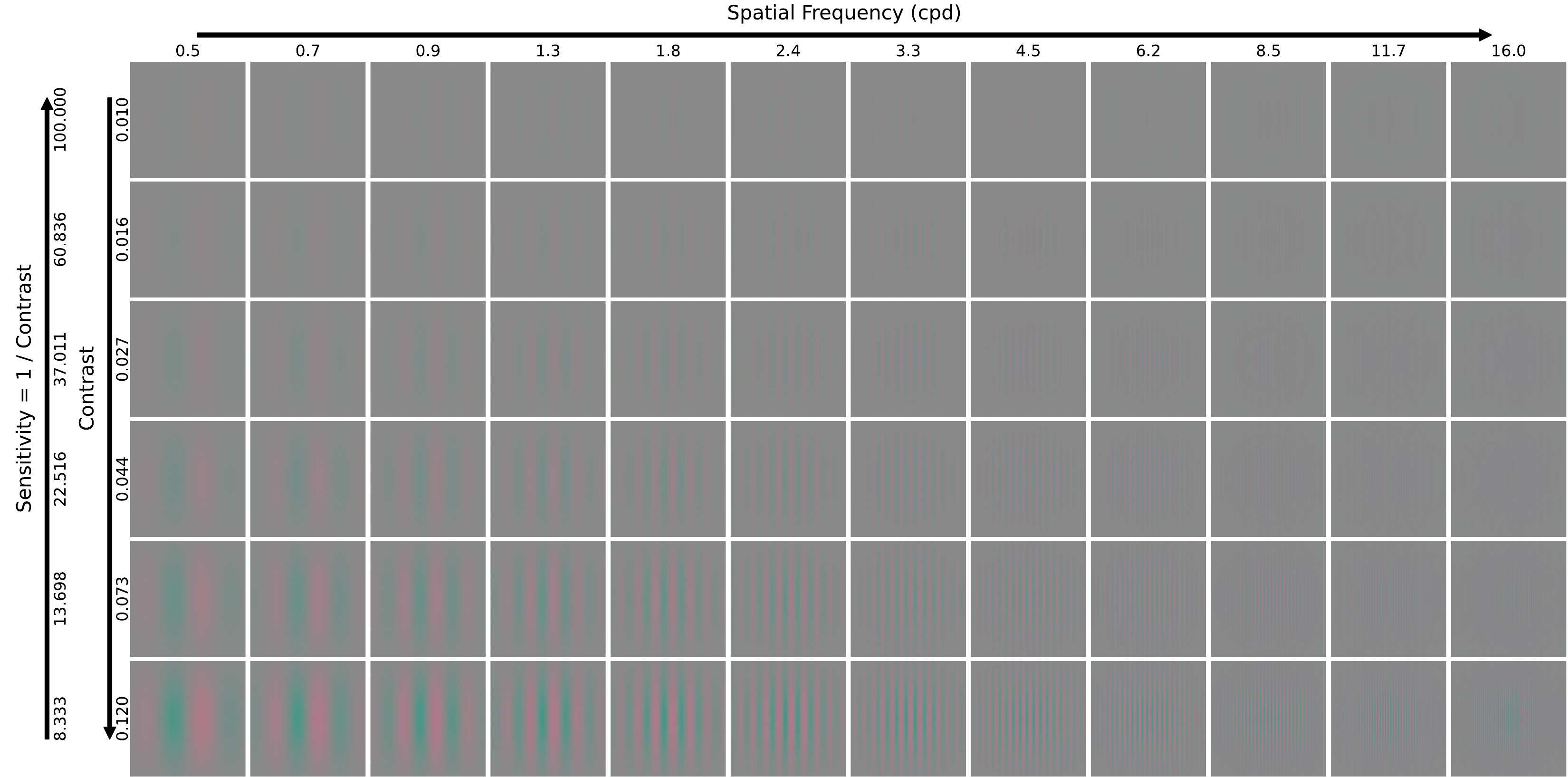}
  \caption{Red-Green (RG) Gabors with different spatial frequencies (x-axis) and contrast (y-axis) used as the test images in the contrast detection tests. Due to gamut limitations, the maximum achievable contrast is capped at 0.2. It was observed that, compared to achromatic Gabor patterns, humans are more sensitive to low frequencies when viewing red-green Gabor patterns.}
  \label{fig:gabor_spf_contrast_rg_sup}
\end{figure*}

\paragraph{Spatial Frequency - Chromatic (RG) - Gabor}
The radius was set to 1$^\circ$, and the background luminance was 100\cdms{}. Test examples are shown in~\figref{gabor_spf_contrast_rg_sup}.

\begin{figure*}[ht]
  \centering
      \includegraphics[width=\linewidth]{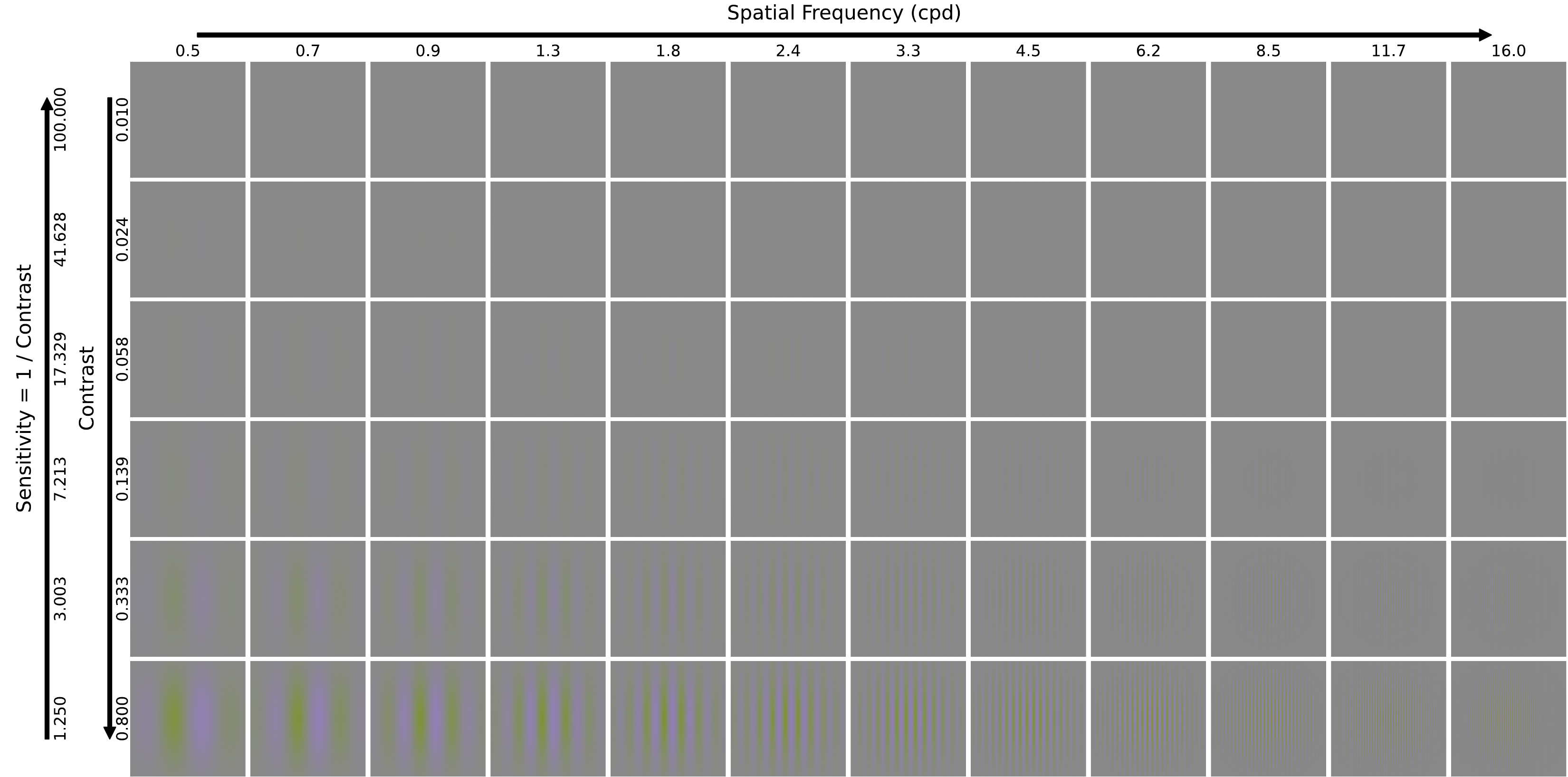}
  \caption{Yellow-Violet (YV) Gabors with different spatial frequencies (x-axis) and contrast (y-axis) used as the test images in the contrast detection tests. Due to gamut limitations, the maximum achievable contrast is capped at 0.2. Similar to RG Gabors, humans are also more sensitive to low-frequency YV Gabors.}
  \label{fig:gabor_spf_contrast_yv_sup}
\end{figure*}

\paragraph{Spatial Frequency - Chromatic (YV) - Gabor}
The radius was set to 1$^\circ$, and the background luminance was 100\csdm. Test examples are shown in~\figref{gabor_spf_contrast_yv_sup}.

\begin{figure*}[ht]
  \centering
      \includegraphics[width=\linewidth]{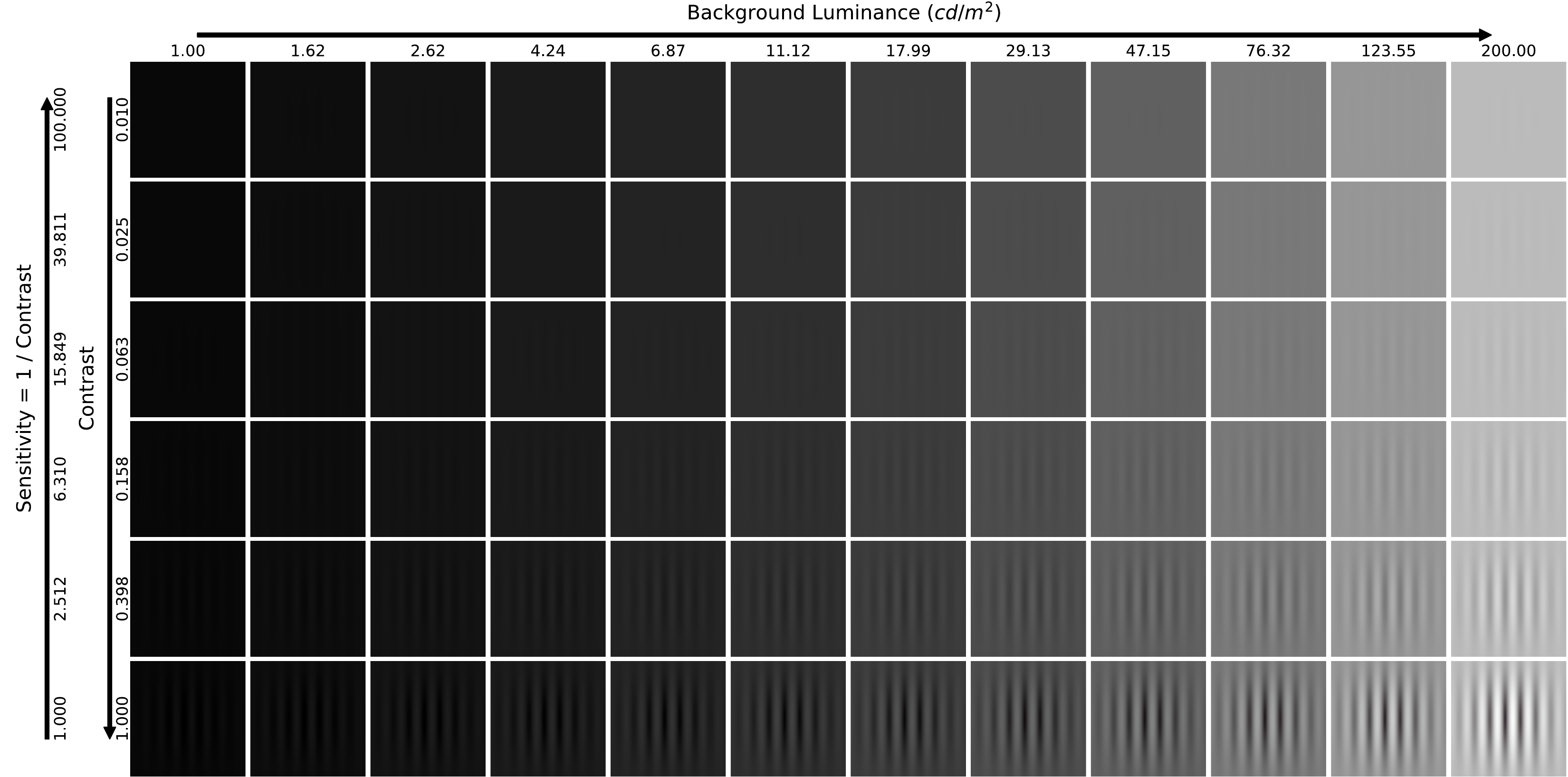}
  \caption{Achromatic Gabor patches with different background luminance (x-axis) and contrasts (y-axis) used as the test images in the contrast detection tests. Note that very low luminance levels cannot be displayed; therefore, a minimum of 1\csdm is used here. In the experiment, this limitation is not present as we use floating-point inputs.}
  \label{fig:gabor_luminance_contrast_sup}
\end{figure*}

\paragraph{Luminance}
The radius was set to 1$^\circ$, and the spatial frequency was 2\,cpd. Test examples are shown in~\figref{gabor_luminance_contrast_sup}.

\begin{figure*}[ht]
  \centering
      \includegraphics[width=\linewidth]{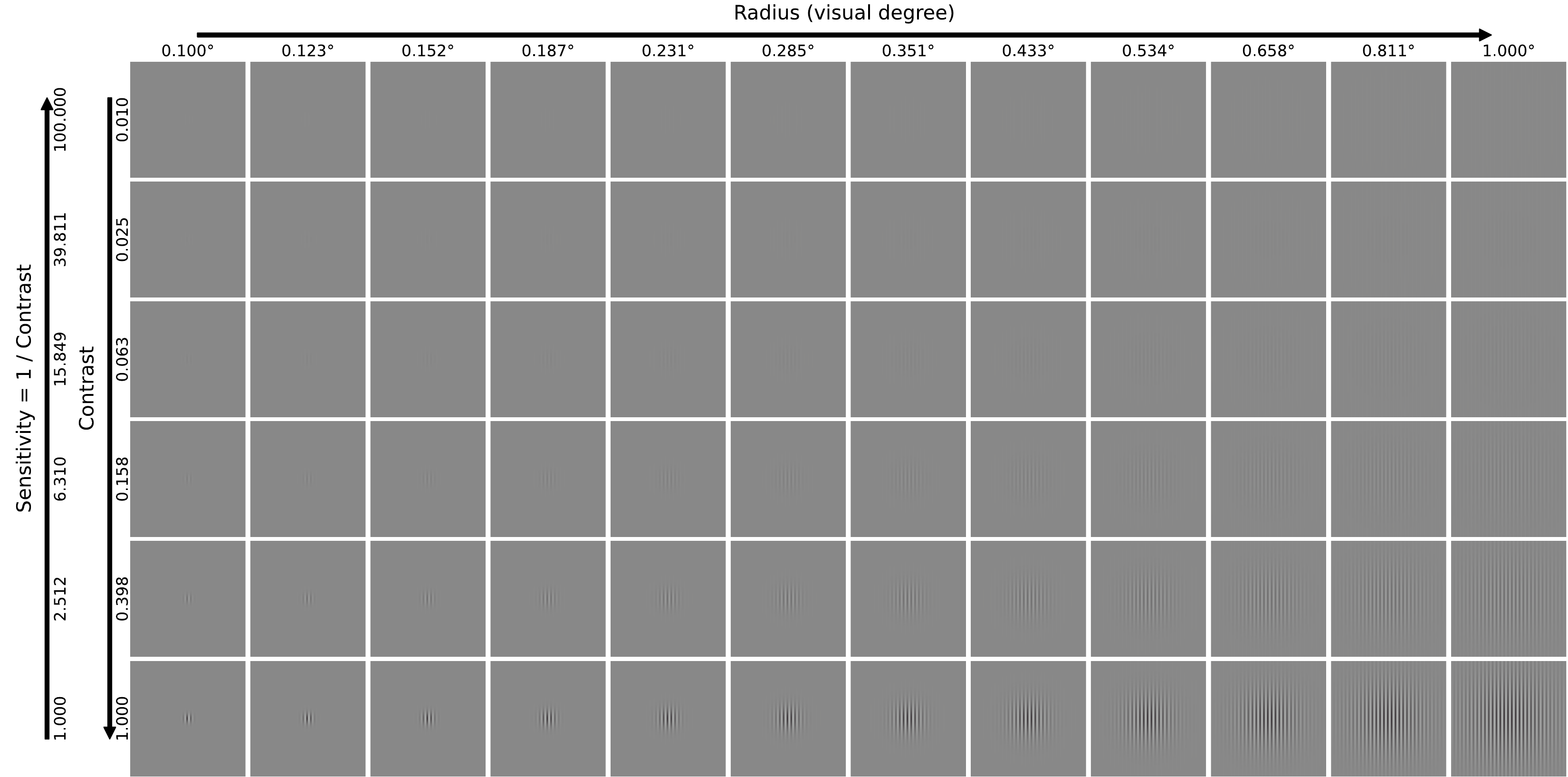}
  \caption{Achromatic Gabor patches with different area (radius) (x-axis) and contrasts (y-axis) used as the test images in the contrast detection tests. In this experiment, we selected a higher spatial frequency (8\,cpd); otherwise, it would be impossible to observe a complete Gabor signal within small areas.}
  \label{fig:gabor_area_contrast_sup}
\end{figure*}

\paragraph{Area}
The background luminance was 100\cdms{}, and the spatial frequency was 8 cycles per degree (cpd). Test examples are shown in~\figref{gabor_area_contrast_sup}.

\subsection{Contrast masking}
\begin{figure*}[ht]
  \centering
      \includegraphics[width=\linewidth]{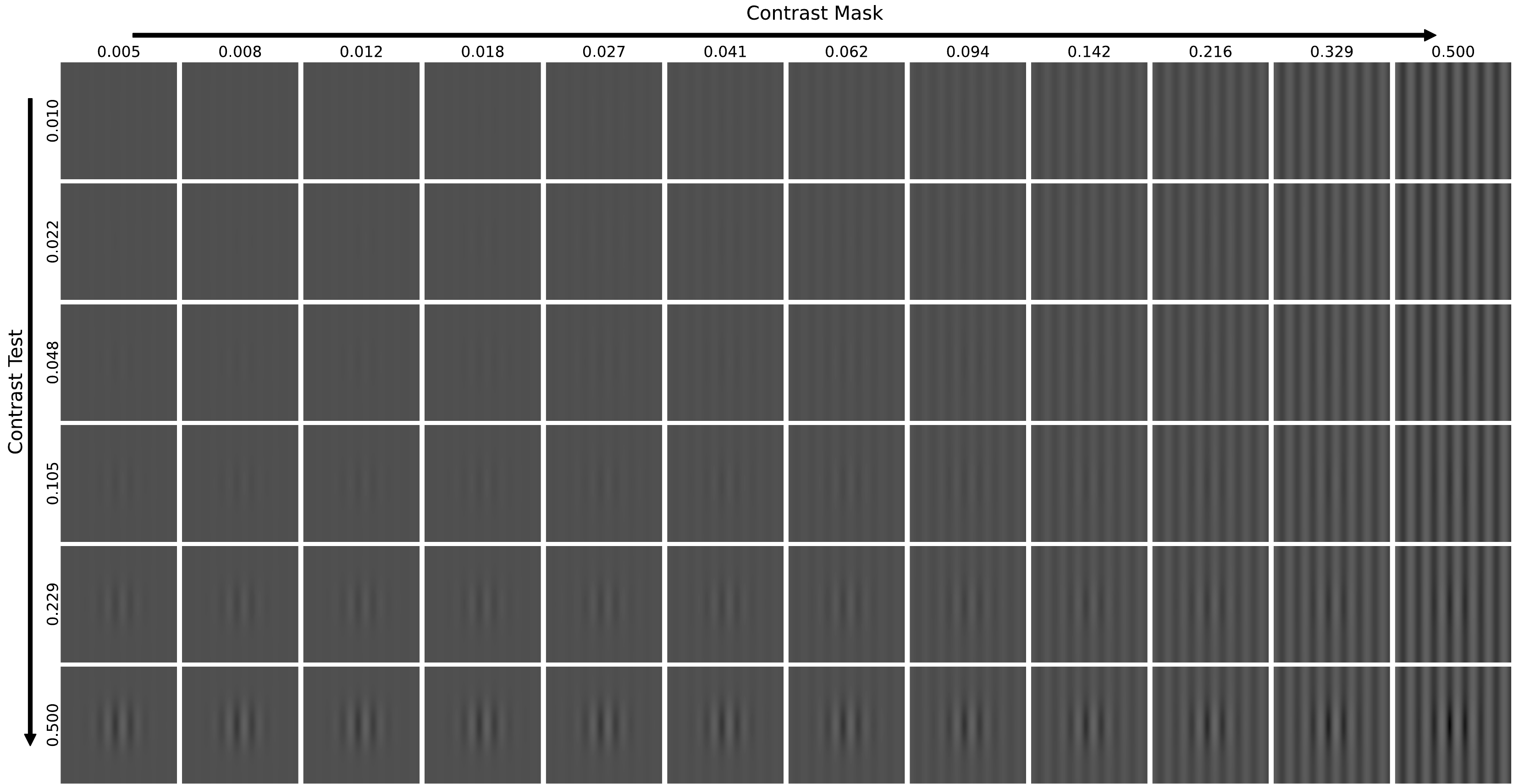}
  \caption{Images from the Phase-Coherent Masking experiment with varying Contrast Mask (x-axis) and Contrast Test (y-axis). The masks are sinusoidal gratings, while the test stimuli are Gabor patterns, set against a background luminance of 32\cdms{}.}
  \label{fig:contrast_masking_gabor_sup}
\end{figure*}

\paragraph{Phase-Coherent Masking}
The test images contained Gabor patches with a spatial frequency of 2\,cpd and a radius of 0.5$^\circ$, while the masks were sinusoidal gratings at the same spatial frequency of 2\,cpd.  The background luminance was 32\cdms{}, following the parameters established in~\cite{foley1994human}. Test examples are shown in~\figref{contrast_masking_gabor_sup}.

\begin{figure*}[ht]
  \centering
      \includegraphics[width=\linewidth]{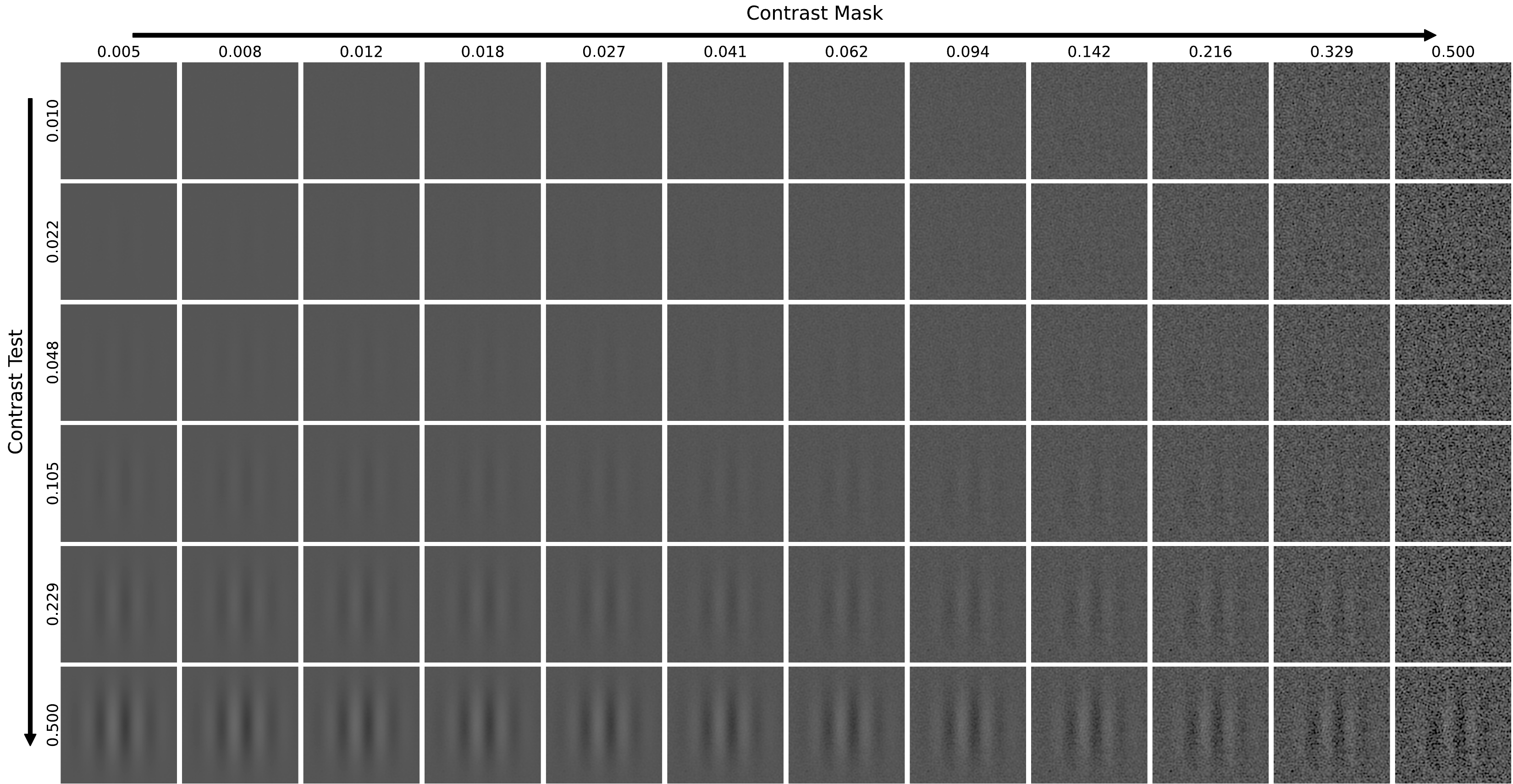}
  \caption{Images from the Phase-Incoherent Masking experiment with varying Contrast Mask (x-axis) and Contrast Test (y-axis). The masks consist of random noise with a frequency spectrum extending up to 12\,cpd, while the test stimuli are Gabor patches, presented against a background luminance of 37\cdms{}.
  }
  \label{fig:contrast_masking_noise_sup}
\end{figure*}

\paragraph{Phase-Incoherent Masking}
The test images contained Gabor patches with a spatial frequency of 1.2\,cpd and a radius of 0.8$^\circ$, and the masks contained random noise with a frequency spectrum extending up to 12\,cpd. The following equations outline the process of generating the noise mask $I_{mask} \in \mathbb{R}^{W \times H}$:

First, Gaussian noise $N(x, y)$ is generated:
\begin{equation}
N(x, y) \sim \mathcal{N}(0, 1), \quad x \in [0, W], \, y \in [0, H]
\label{eq:PIN_1}.
\end{equation}
Next, a two-dimensional Fast Fourier Transform (FFT) is applied to obtain the frequency domain representation $N_f(u, v)$:
\begin{equation}
N_f(u, v) = \mathcal{F}\{N(x, y)\} , \quad u \in [0, W], \, v \in [0, H]
\label{eq:PIN2}.
\end{equation}
Subsequently, frequency filtering is performed:
\begin{equation}
N_f^{\text{filtered}}(u, v) = 
\begin{cases} 
N_f(u, v), & \rho(u, v) \leq 12 \, \text{cpd} \\
0, & \rho(u, v) > 12 \, \text{cpd}
\end{cases}
\label{eq:PIN_3},
\end{equation}

\begin{equation}
\rho(u, v) = \sqrt{(K_u)^2 + (K_v)^2}
\label{eq:PIN_4},
\end{equation}
\begin{equation}
K_u = 2\,\rho_{\text{nyquist}}\left(\text{mod}\left(\frac{1}{2} + \frac{u}{W}, 1\right) - \frac{1}{2}\right)
\label{eq:PIN_5},
\end{equation}
\begin{equation}
K_v = 2\,\rho_{\text{nyquist}}\left(\text{mod}\left(\frac{1}{2} + \frac{v}{H}, 1\right) - \frac{1}{2}\right)
\label{eq:PIN_6},
\end{equation}
where $\rho_{\text{nyquist}} = \frac{\text{ppd}}{2}$. The noise in the spatial domain is then obtained using the inverse Fourier transform:
\begin{equation}
N_{\text{bp}}(x, y) = \mathcal{F}^{-1}\{N_f^{\text{filtered}}(u, v)\}
\label{eq:PIN_7}.
\end{equation}
Finally, the noise mask $I_{mask}$ is generated:
\begin{equation}
I_{mask}(x, y) = L_b \,\left(1 + c_{\text{mask}} \,\frac{N_{\text{bp}}(x, y)}{\sigma_{N_{\text{bp}}}}\right)
\label{eq:PIN_8},
\end{equation}
where $c_{\text{mask}}$ is the mask contrast, $\sigma_{N_{\text{bp}}}$ represents the standard deviation of $N_{\text{bp}}$. The background luminance $L_b$ was 37\cdms{}, consistent with the conditions in~\cite{gegenfurtner1992contrast}. Test examples are shown in~\figref{contrast_masking_noise_sup}.

\subsection{Supra-threshold contrast matching}
\label{sec:subsec_supra_CM}
\begin{figure*}[ht]
  \centering
      \includegraphics[width=\linewidth]{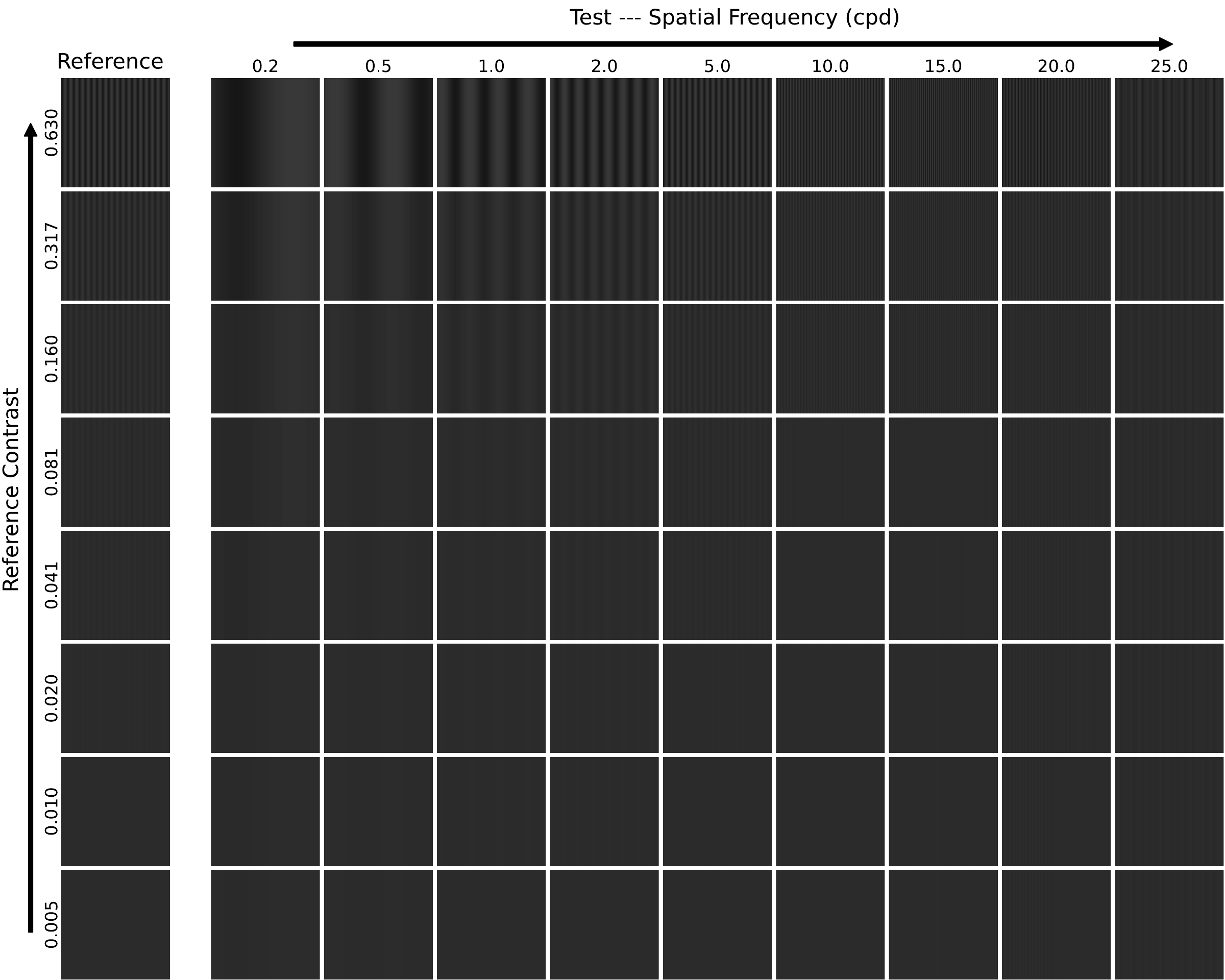}
  \caption{Example images from the Contrast Matching experiment. The first column on the left displays the references, which are sinusoidal gratings at 5 cycles per degree (cpd) with varying reference contrasts. The remaining images on the right are tests with different spatial frequencies matched to the references. Note that the contrast levels of the references, as well as the contrasts and spatial frequencies of the tests, are based on the experimental results in~\cite{georgeson1975contrast}. Following the experimental conditions outlined in~\cite{georgeson1975contrast}, the background luminance is set to 10\cdms{}.
  }
  \label{fig:contrast_matching_sup}
\end{figure*}

We followed the experimental setup from~\cite{georgeson1975contrast}, where the reference was a sinusoidal grating with a spatial frequency of 5\,cpd and a luminance of 10\cdms{}, presented at eight distinct contrast levels $c_{r}$. The test stimulus was also a sinusoidal grating with a luminance of 10\cdms{}, but presented at various spatial frequencies $\rho_{t}$. Examples are shown in~\figref{contrast_matching_sup}.


\begin{figure*}[t]
  \centering
  \includegraphics[width=\linewidth]{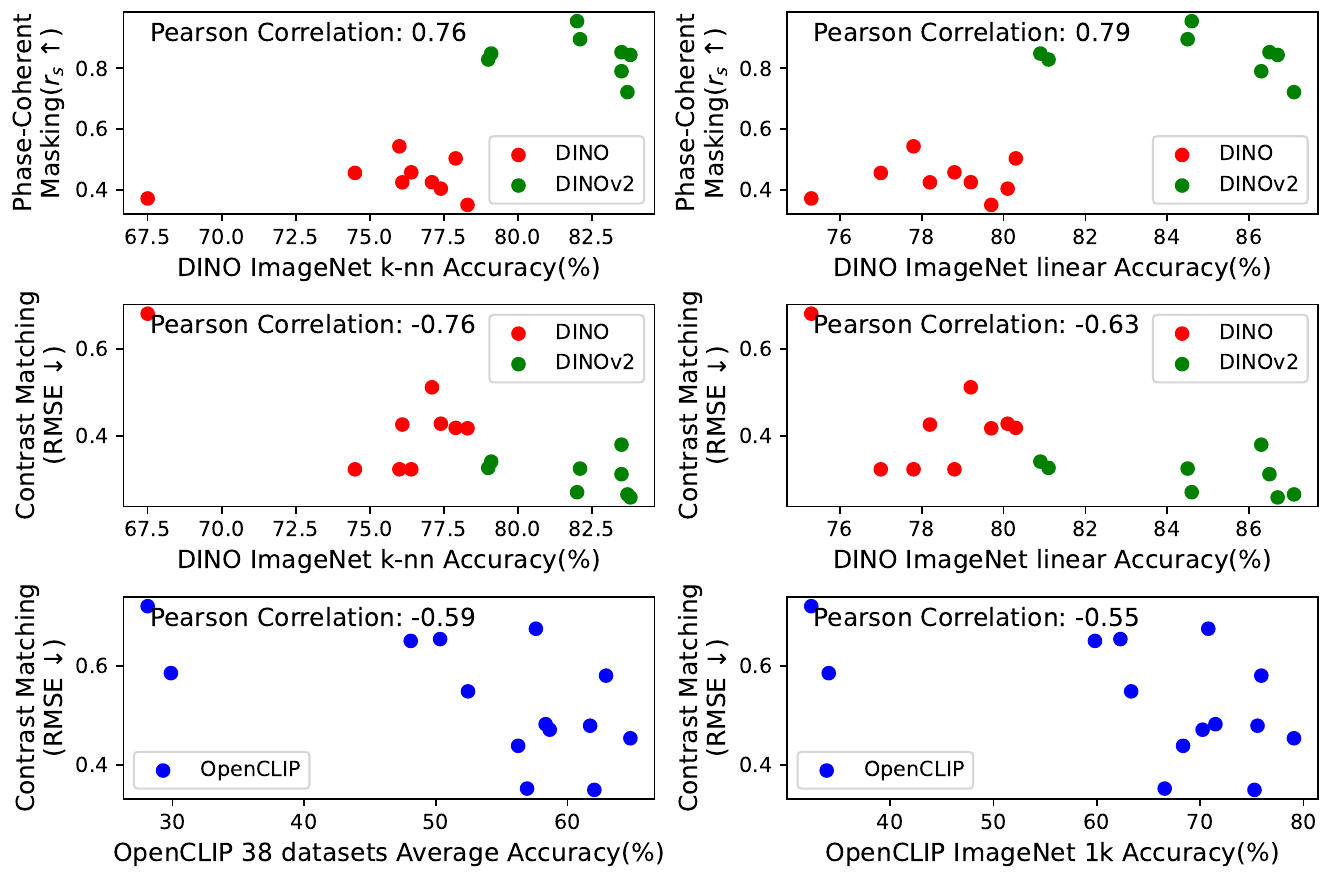}
  \caption{The performance of DINO/DINOv2/OpenCLIP on their classification tasks (from their GitHub repo) shows a potential correlation with the alignment scores in our masking/matching tests.}
  \label{fig:rebuttal}
  % \vspace{-10pt}
\end{figure*}

\section{Practical implications}
We checked whether our alignment scores (Fig. 6 in the paper) can indicate how well a model can perform on computer vision tasks. In Figure~\ref{fig:rebuttal}, we show scatter plots of the alignment scores and different performance indicators for DINO, DINOv2, and OpenCLIP (data were not available for other models). The correlations (absolute value 0.55–0.8) suggest that good alignment with the contrast masking/matching characteristic can improve model's performance. Such results were consistent for the alignment of contrast masking and contrast matching, less so for detection (as expected). We did not find a strong correlation between alignment scores and the parameters of the model architecture (model size, number of parameters) or computational GFlops. We hope that future work can provide stronger evidence for the benefits of model-HVS alignment and spark interest in using low-level human vision models to introduce invariances or constraints into the training of the foundation models (via architectural changes, loss functions, or data augmentation).

\section{Model alignment scores}
Section 3.2 in the main text briefly describes the computation of Spearman rank-order correlation coefficients for model alignment scores. This section provides further details and formulas.

Specifically, for each contour plot, $N$ points were selected along the x-axis $X_{1 \ldots N}$, where $X$ represents dimensions such as area, luminance, or mask contrast. Based on the predictions of castleCSF, we obtain the ground truth $Y_{1 \ldots N}$, where $Y$ represents sensitivity in contrast detection and test contrast in contrast masking.

We then scaled each $Y_j (j = 1\ldots N)$ by multipliers $m_i (i = 1\ldots M)$:
\begin{equation}
m_i = 10^{\log_{10}(0.5) + \frac{i - 1}{M - 1} \cdot \log_{10}\left(\frac{2}{0.5}\right)}
\label{eq:multiplier_1},
\end{equation}
\begin{equation}
Y'_{ij} = m_iY_j
\label{eq:multiplier_2},
\end{equation}
producing $Y'_{1 \ldots NM
}$ and their respective $S_{1 \ldots NM
}$ ($S_\ind{ac}$)\footnote{Specifically, for these \(NM\) conditions, a test and reference signal pair is generated for each condition ($X_j, Y'_{ij}$), and \(S_{ac}\) is computed using the method described in the main text.}. Given that psychometric functions near the threshold typically exhibit uniform shapes across all conditions in psychophysical experiments, we hypothesized that the trend of scaled scores would remain consistent across all $Y_{1..N}$. The Spearman's rank correlation coefficient $r_s$ was calculated as the similarity metric:
\begin{equation}
r_s = \frac{\operatorname{cov}(\mathrm{R}(m_{1 \ldots N M}), \mathrm{R}(S_{1 \ldots NM}))}{\sigma_{\mathrm{R}(m_{1 \ldots N M})} \sigma_{\mathrm{R}(S_{1 \ldots NM})}}
\label{eq:spearman},
\end{equation}
where $m_{1 \ldots N M}=\bigcup_{k=1}^{N} m_{1 \ldots M}$, $\mathrm{R}(*)$ denotes ranked data, $\operatorname{cov}(*)$ represents covariance, and $\sigma(*)$ stands for standard deviation. For all models and tests, higher $r_s$ (closer to 1) reflects a greater model alignment. In the contrast detection experiment, $N=20, M=10$. In the contrast masking experiment, $M=10$ and $N$ is equal to the number of human data points.

In the main text, we presented the experimental results for all models in the form of bar charts. To provide higher decimal precision, the results are presented in Table~\ref{tab:results}.

\begin{table*}[t]
\begin{center}
\caption{The model alignment scores for all 45 models across nine test types. Spearman's rank correlation coefficient \( r_s \) is used as the evaluation metric for the contrast detection and contrast masking experiments, with higher values (approaching 1) indicating greater similarity between the model and the human visual system. For the contrast matching experiment, Root Mean Square Error (RMSE) is employed as the metric, where lower values (approaching 0) signify a closer match to the human visual system. For each model series, its best score on each test has been highlighted in bold.}
\label{tab:results}
\scriptsize
\renewcommand{\arraystretch}{1.3}
\setlength{\tabcolsep}{0.5mm}{\scalebox{1.06}{
\begin{tabular}{c|c|c|c|c|c|c|c|c|c|c|c}
\toprule
Models & Architecture & \begin{tabular}[c]{@{}c@{}}Training\\      dataset\end{tabular} & \begin{tabular}[c]{@{}c@{}}Spatial Freq.\\      Gabor Ach.\\      $r_s$↑\end{tabular} & \begin{tabular}[c]{@{}c@{}}Spatial Freq.\\      Noise Ach.\\      $r_s$↑\end{tabular} & \begin{tabular}[c]{@{}c@{}}Spatial Freq.\\      Gabor RG\\      $r_s$↑\end{tabular} & \begin{tabular}[c]{@{}c@{}}Spatial Freq.\\      Gabor YV\\      $r_s$↑\end{tabular} & \begin{tabular}[c]{@{}c@{}}Luminance\\      Gabor Ach.\\      $r_s$↑\end{tabular} & \begin{tabular}[c]{@{}c@{}}Area\\      Gabor Ach.\\      $r_s$↑\end{tabular} & \begin{tabular}[c]{@{}c@{}}Phase\\      Coherent\\      Masking\\      $r_s$↑\end{tabular} & \begin{tabular}[c]{@{}c@{}}Phase\\      Incoherent\\      Masking\\      $r_s$↑\end{tabular} & \begin{tabular}[c]{@{}c@{}}Contrast\\      Matching\\      RMSE↓\end{tabular} \\ \hline\hline
No Encoder & - & - & 0.4688 & 0.4594 & 0.5235 & 0.6582 & 0.4188 & 0.8981 & 0.5057 & 0.6746 & 0.2657 \\ \hline\hline
\multirow{9}{*}{DINO} & ResNet-50 & ImageNet & 0.3428 & 0.2506 & 0.2795 & 0.3359 & \textbf{0.5623} & 0.9610 & 0.3716 & 0.\textbf{8913} & 0.6803 \\ \cline{2-12} 
 & ViT-S/16 & ImageNet & 0.4769 & \textbf{0.4951} & 0.5316 & 0.4517 & 0.4260 & \textbf{0.9686} & 0.4556 & 0.8844 & \textbf{0.3238}\\ \cline{2-12} 
 & ViT-S/8 & ImageNet & 0.4129 & 0.4248 & 0.4211 & 0.4114 & 0.4048 & 0.9358 & 0.3504 & 0.6262 & 0.4178 \\ \cline{2-12} 
 & ViT-B/16 & ImageNet & 0.5281 & 0.4883 & \textbf{0.6148} & \textbf{0.4699} & 0.4554 & 0.9584 & 0.4246 & 0.6351 & 0.4264 \\ \cline{2-12} 
 & ViT-B/8 & ImageNet & 0.4213 & 0.4446 & 0.4398 & 0.4184 & 0.3655 & 0.8929 & 0.4039 & 0.4761 & 0.4283 \\ \cline{2-12} 
 & Xcit-S-12/16 & ImageNet & \textbf{0.5721} & 0.4783 & 0.5312 & 0.4315 & 0.4436 & 0.9418 & \textbf{0.5431} & 0.7120 & 0.3239 \\ \cline{2-12} 
 & Xcit-S-12/8 & ImageNet & 0.4337 & 0.3961 & 0.3862 & 0.3064 & 0.4373 & 0.8409 & 0.4250 & 0.7618 & 0.5117 \\ \cline{2-12} 
 & Xcit-M-24/16 & ImageNet & 0.5424 & 0.4848 & 0.4651 & 0.4330 & 0.4916 & 0.9408 & 0.4576 & 0.6458 & 0.3238 \\ \cline{2-12} 
 & Xcit-M-24/8 & ImageNet & 0.4852 & 0.3848 & 0.3742 & 0.4030 & 0.5172 & 0.8149 & 0.5034 & 0.8028 & 0.4187 \\ \hline
\multirow{8}{*}{DINOv2} & ViT-S/14 & LVD-142M & 0.3976 & 0.4348 & 0.4114 & 0.5207 & 0.4865 & 0.7071 & 0.8288 & 0.9593 & 0.3271 \\ \cline{2-12} 
 & ViT-B/14 & LVD-142M & 0.4286 & 0.5032 & 0.4898 & 0.7148 & 0.5580 & 0.9216 & 0.8955 & 0.8514 & 0.3255 \\ \cline{2-12} 
 & ViT-L/14 & LVD-142M & \textbf{0.5256} & 0.4843 & 0.5152 & 0.4934 & 0.5363 & \textbf{0.9493} & 0.7902 & 0.7968 & 0.3803 \\ \cline{2-12} 
 & ViT-g/14 & LVD-142M & 0.4304 & 0.5198 & \textbf{0.5277} & 0.6687 & 0.4935 & 0.7856 & 0.8530 & 0.8846 & 0.3127 \\ \cline{2-12} 
 & ViT-S/14 + reg & LVD-142M & 0.4508 & 0.4484 & 0.4254 & 0.4252 & 0.5201 & 0.6942 & 0.8484 & 0.9254 & 0.3415 \\ \cline{2-12} 
 & ViT-B/14 + reg & LVD-142M & 0.4408 & 0.5180 & 0.4837 & 0.6488 & \textbf{0.5705} & 0.4752 & \textbf{0.9549} & 0.9539 & 0.2714 \\ \cline{2-12} 
 & ViT-L/14 + reg & LVD-142M & 0.4423 & 0.5645 & 0.4799 & \textbf{0.7996} & 0.5167 & 0.5289 & 0.8439 & \textbf{0.9814} & \textbf{0.2595} \\ \cline{2-12} 
 & ViT-g/14 + reg & LVD-142M & 0.4351 & \textbf{0.5732} & 0.4518 & 0.6670 & 0.5168 & 0.5600 & 0.7214 & 0.9295 & 0.2663 \\ \hline
\multirow{14}{*}{OpenCLIP} & ResNet-50 & OpenAI & 0.3499 & 0.2903 & 0.2972 & 0.3310 & \textbf{0.5703} & 0.6855 & 0.4981 & 0.7349 & 0.6505 \\ \cline{2-12} 
 & ResNet-50 & YFCC-15M & 0.3604 & 0.2562 & 0.3206 & 0.2729 & 0.5629 & \textbf{0.9671} & 0.4506 & 0.6802 & 0.7208 \\ \cline{2-12} 
 & ResNet-101 & OpenAI & 0.4130 & 0.3001 & 0.3403 & 0.3530 & 0.5070 & 0.9568 & 0.4031 & 0.5697 & 0.6542 \\ \cline{2-12} 
 & ResNet-101 & YFCC-15M & 0.3414 & 0.2275 & 0.3620 & 0.3479 & 0.4569 & 0.9046 & 0.5136 & 0.8588 & 0.5853 \\ \cline{2-12} 
 & ConvNext-B-w & LAION-2B & 0.3957 & 0.2787 & 0.3772 & 0.4233 & 0.4802 & 0.9590 & 0.3461 & 0.6583 & 0.6752 \\ \cline{2-12} 
 & ConvNext-B-w & LAION-2B+ & 0.4649 & 0.3487 & 0.4458 & 0.5648 & 0.3854 & 0.7026 & 0.4887 & 0.5971 & 0.4820 \\ \cline{2-12} 
 & ConvNext-L-d & LAION-2B+ & 0.3419 & 0.2835 & 0.3911 & 0.6443 & 0.1530 & 0.7690 & 0.4919 & 0.4983 & 0.5803 \\ \cline{2-12} 
 & ConvNext-XXL & LAION-2B+ & 0.4136 & 0.3642 & 0.3890 & 0.6212 & 0.1112 & 0.8422 & 0.4699 & 0.4586 & 0.4535 \\ \cline{2-12} 
 & ViT-B/32 & OpenAI & 0.4132 & 0.5696 & 0.3748 & 0.7197 & 0.1772 & 0.9434 & 0.4047 & 0.7736 & 0.5484 \\ \cline{2-12} 
 & ViT-B/32 & LAION-2B & 0.5108 & \textbf{0.6556} & 0.3146 & 0.4108 & 0.3063 & 0.8673 & \textbf{0.8837} & \textbf{0.9620} & 0.3517 \\ \cline{2-12} 
 & ViT-B/16 & OpenAI & 0.4798 & 0.5429 & 0.4138 & 0.7141 & 0.2655 & 0.7887 & 0.4321 & 0.6451 & 0.4382 \\ \cline{2-12} 
 & ViT-B/16 & LAION-2B & 0.4654 & 0.5740 & 0.4144 & 0.6922 & 0.4242 & 0.7869 & 0.5235 & 0.7725 & 0.4709 \\ \cline{2-12} 
 & ViT-L/14 & OpenAI & 0.4625 & 0.5026 & 0.4357 & 0.6229 & 0.4585 & 0.8892 & 0.5496 & 0.7050 & 0.4789 \\ \cline{2-12} 
 & ViT-L/14 & LAION-2B & \textbf{0.5917} & 0.5240 & \textbf{0.4801} & \textbf{0.7338} & 0.3084 & 0.6430 & 0.7678 & 0.8799 & \textbf{0.3490} \\ \hline
\multirow{3}{*}{SAM} & ViT-B-SAM & SA-1B & 0.3545 & \textbf{0.3287} & \textbf{0.3577} & 0.3617 & 0.3074 & \textbf{0.9714} & \textbf{0.4144} & 0.4353 & 0.5877 \\ \cline{2-12} 
 & ViT-L-SAM & SA-1B & 0.3061 & 0.2769 & 0.3160 & 0.3140 & 0.3090 & 0.9598 & 0.4077 & 0.3787 & 0.6354 \\ \cline{2-12} 
 & ViT-H-SAM & SA-1B & \textbf{0.3651} & 0.3234 & 0.3316 & \textbf{0.3863} & \textbf{0.5243} & 0.9533 & 0.3983 & \textbf{0.5105} & \textbf{0.5489} \\ \hline
\multirow{4}{*}{SAM-2} & SAM2.1-hiera-tiny & SA-V & 0.4058 & 0.3483 & 0.4315 & 0.4693 & 0.4966 & \textbf{0.9660} & 0.4137 & 0.4646 & 0.4805 \\ \cline{2-12} 
 & SAM2.1-hiera-S & SA-V & \textbf{0.4544} & \textbf{0.3705} & 0.3936 & 0.4608 & 0.5389 & 0.9533 & 0.398 & 0.4941 & \textbf{0.4472} \\ \cline{2-12} 
 & SAM2.1-hiera-B+ & SA-V & 0.3728 & 0.3195 & \textbf{0.4949} & 0.4993 & 0.5396 & 0.9296 & 0.4062 & 0.4882 & 0.4852 \\ \cline{2-12} 
 & SAM2.1-hiera-L & SA-V & 0.3872 & 0.3259 & 0.3695 & \textbf{0.4935} & \textbf{0.5631} & 0.9431 & \textbf{0.4613} & \textbf{0.7361} & 0.4686 \\ \hline
\multirow{3}{*}{MAE} & ViT-B-MAE & ImageNet & \textbf{0.4471} & \textbf{0.4903} & \textbf{0.4410} & \textbf{0.5008} & 0.5812 & \textbf{0.9036} & 0.5446 & \textbf{0.7277} & \textbf{0.4223} \\ \cline{2-12} 
 & ViT-L-MAE & ImageNet & 0.4284 & 0.4560 & 0.4126 & 0.4803 & 0.5647 & 0.8874 & \textbf{0.6849} & 0.7043 & 0.4344 \\ \cline{2-12} 
 & ViT-H-MAE & ImageNet & 0.4250 & 0.3969 & 0.3995 & 0.4737 & \textbf{0.6335} & 0.6466 & 0.5003 & 0.6088 & 0.4964 \\ \hline
\multirow{2}{*}{SD-VAE} & SD-v1-5 & LAION-5B & \textbf{0.3527} & \textbf{0.4226} & \textbf{0.8447} & \textbf{0.8051} & \textbf{0.4993} & \textbf{0.8402} & \textbf{0.5394} & \textbf{0.5579} & \textbf{0.4177} \\ \cline{2-12} 
 & SD-xl-base-1.0 & LAION-5B & 0.2465 & 0.1662 & 0.3811 & 0.3132 & 0.4273 & 0.3561 & 0.4996 & 0.4962 & 0.6727 \\ \hline\hline
ColorVideoVDP & HVS-based & XR-DAVID+ & 0.5545 & 0.7817 & 0.7455 & 0.9339 & 0.9020 & 0.8937 & 0.7418 & 0.7626 & 0.2604 \\ \bottomrule
\end{tabular}

}}
\end{center}
\end{table*}

{
    \small
    \bibliographystyle{ieeenat_fullname}
    \bibliography{LVM_sythestic_test}
}